\documentclass[12pt,a4paper]{article}
\usepackage[pass]{geometry}

\usepackage{fullpage}
\usepackage{authblk}
\usepackage[english]{babel}
\usepackage[utf8x]{inputenc}
\usepackage{amsmath}
\usepackage{amssymb}
\usepackage{graphicx}
\graphicspath{ {images/} }
\usepackage{longtable}
\usepackage{multirow}
\usepackage{hyperref}
\usepackage{boldline} 
\usepackage{subcaption}
\usepackage{comment}
\usepackage{mathtools}
\usepackage[table,xcdraw]{xcolor}
\let\oldFootnote\footnote
\newcommand\nextToken\relax

\renewcommand\footnote[1]{%
    \oldFootnote{#1}\futurelet\nextToken\isFootnote}

\newcommand\isFootnote{%
    \ifx\footnote\nextToken\textsuperscript{,}\fi}

\providecommand{\keywords}[1]{\textbf{\textit{Index terms---}} #1}

\newcommand\newsubcap[1]{\phantomcaption%
       \caption*{\thefigure(\thesubfigure)}}

\title{EDITH : \textbf{E}CG biometrics aided by \textbf{D}eep learning for reliable \textbf{I}ndividual au\textbf{TH}entication}

\author[1]{Nabil Ibtehaz}
\author[2,*]{Muhammad~E.~H.~Chowdhury}
\author[2]{Amith~Khandakar}
\author[2]{Serkan~Kiranyaz}
\author[1]{M. Sohel Rahman}
\author[2]{Anas~Tahir}
\author[2]{Yazan~Qiblawey}
\author[3]{Tawsifur~Rahman}

\affil[1]{Department of Computer Science and Engineering, Bangladesh University of Engineering and Technology, Dhaka-1205, Bangladesh}
\affil[2]{Department of Electrical Engineering, Qatar University, Doha-2713, Qatar}
\affil[3]{Department of Biomedical Physics \& Technology, University of Dhaka, Dhaka-1000, Bangladesh}
\affil[*]{Corresponding author : mchowdhury@qu.edu.qa}

\begin{document}

\maketitle

\begin{abstract}

In recent years, physiological signal-based authentication has shown great promises, for its inherent robustness against forgery. Electrocardiogram (ECG) signal, being the most widely studied biosignal, has also received the highest level of attention in this regard. It has been proven with numerous studies that by analyzing ECG signals from different persons, it is possible to identify them, with acceptable accuracy. In this work, we present, EDITH, a deep learning-based framework for ECG biometrics authentication system. Moreover, we hypothesize and demonstrate that Siamese architectures can be used over typical distance metrics for improved performance. We have evaluated EDITH using 4 commonly used datasets and outperformed the prior works using a fewer number of beats. EDITH performs competitively using just a single heartbeat (96$\sim$99.75\% accuracy) and can be further enhanced by fusing multiple beats (100\% accuracy from 3 to 6 beats). Furthermore, the proposed Siamese architecture manages to reduce the identity verification Equal Error Rate (EER) to 1.29 \%. A limited case study of EDITH with real-world experimental data also suggests its potential as a practical authentication system. 
\end{abstract}


\footnotetext{© 2021 IEEE.  Personal use of this material is permitted.  Permission from IEEE must be obtained for all other uses, in any current or future media, including reprinting/republishing this material for advertising or promotional purposes, creating new collective works, for resale or redistribution to servers or lists, or reuse of any copyrighted component of this work in other works.
}

%\begin{footnote}© 2021 IEEE.  Personal use of this material is permitted.  Permission from IEEE must be obtained for all other uses, in any current or future media, including reprinting/republishing this material for advertising or promotional purposes, creating new collective works, for resale or redistribution to servers or lists, or reuse of any copyrighted component of this work in other works.\end{footnote}

\section{Introduction}
 
Since the dawn of civilization, the human being has been storing valuable objects ensuring secrecy and security. Mechanisms such as keys or combination locks have been used to serve such purposes for centuries. However, with the development of information technology, our valuable possessions (e.g., bank accounts) have mostly been shifted to the digital domain. Besides, various personalized services and amenities have also received similar technological transformation. The concern of security and secrecy still applies; thus various measures are adopted to allow only authorized access. Analogous to the concepts of keys and combination locks, in the digital domain, tokens and passwords have been being used for decades. But they come with the risk of being lost or stolen, bringing in the peril of malicious access \cite{unar2014review}.

Therefore, a multitude of research works have been conducted for decades to develop resilient authentication systems. There exist a lot works in the literature that focused on behavioral modalities, such as voice \cite{kim2010person}, keystroke \cite{balagani2011discriminability},  gait \cite{gafurov2007gait} etc. However, the behavioral authentication systems are often cumbersome and unreliable. This has made biometric authentication systems more preferred, which include fingerprint \cite{clancy2003secure}, retina \cite{condurache2012robust}, facial recognition \cite{lee2003new} etc. Despite being widely adopted, these methods also face a serious issue of forgery attacks \cite{jain2012biometric}.

In recent years, physiological signals have also demonstrated great potential in authenticating individuals. Physiological signals are in fact more preferable to other means of authentication as they are hard to counterfeit and also the individuals need to present themselves to record such signals \cite{labati2019deep}. Among various physiological signals electrocardiogram (ECG) \cite{biel2001ecg}, electroencephalography (EEG) \cite{maiorana2019eeg} and photoplethysmography (PPG) \cite{yadav2018evaluation} are the most prominent. Compared to most physiological signals, PPG and ECG signals can be obtained relatively easily, for example, using simple finger sensors \cite{evans2017feasibility}. Furthermore, ECG signals are more resilient against noises, compared to PPG or EEG, which has made ECG more preferable as a biometric system \cite{israel2005ecg}. As a result ECG bands \cite{kang2016ecg}, low power circuits \cite{yin2016low}  and light-weight algorithms \cite{chun2016ecg} have been developed for ECG based authentication systems. Furthermore, with the recent introduction of ECG sensors in commercial products like the Apple Watch and Samsung Galaxy Watch, the possibility of widespread adoption of ECG biometrics has opened \cite{kim2019cancelable}.

In this paper, we propose EDITH, which is a deep learning-based ECG biometric authentication system, EDITH. EDITH comprises a novel convolutional network architecture based on multiresolution analysis, employing MultiRes blocks and Spatial Pyramid Pooling in coherent conjunction. To satisfy the reliance on accurate R-peak detection, we develop a deep learning-based R-peak detector. Furthermore, we propose a novel Siamese architecture incorporating both Euclidean and Cosine similarity distance measures. EDITH has been evaluated on 4 benchmark datasets and have consistently outperformed the prior works.

\section{Related Works}

The pioneering work of Biel et al. \cite{biel2001ecg} deserves the most credit for establishing the potential of ECG signals for human identification. Ever since this pilot study, ECG, as a biometric has drawn widespread interest, resulting in a plethora of novel approaches\cite{pinto2018evolution}. The different ECG based authentication systems can broadly be divided into two classes, fiducial and non-fiducial.

The fiducial methods rely on extracting various fiducial landmarks from the ECG signal. For such approaches, different sub-waves like Q, R, S, P, T are extracted, and features related to various amplitude, difference, or delay terms are computed \cite{zhang2006new,shen2010implementation,venkatesh2010human}. However, these methods are often criticized for over-reliance on the fiducial points, slight misdetection of which results in an erroneous identification \cite{pinto2018evolution}. The inability of working with noisy or variable heart-beats seriously limits these methods. Moreover, for satisfactory performance, manual intervention is often required, which further diminishes their utility \cite{chu2019ecg}.

To overcome the limitations of the fiducial methods, subsequently, non-fiducial approaches were investigated. These methods analyze the entire or a segment of the signal to extract several morphological features \cite{lourencco2011unveiling}. Among the diverse exploration of non-fiducial methods, the use of Kalman filter \cite{ting2010ecg}, Fourier or Wavelet transformation \cite{saechia2005human}, and generalized S-transformation \cite{zhao2018ecg} are the most notable ones.

Recently, several deep learning-based methods have also been proposed for ECG biometrics. Such methods have demonstrated much greater accuracy in identification compared to the traditional approaches. Although such methods often rely on the detection of R-peaks for proper alignment of the signals, they still circumvent the flaws of the fiducial approaches. Since R-peak is the highest and sharpest point of ECG and the QRS complex is less likely to be affected by physical movement or emotion, they are popularly used for ECG signal alignment \cite{labati2019deep}. Convolutional networks, ranging from vanilla architectures \cite{labati2019deep}, multi-resolutional models \cite{zhang2017heartid,chu2019ecg}, residual models \cite{ihsanto2020fast} have all demonstrated superior accuracy in this context. Similarly, recurrent network models such as Long Short-Term Memory (LSTM) \cite{salloum2017ecg} or Gated Recurrent Unit (GRU) \cite{lynn2019deep} have also been successfully applied for precise individual authentication.

\section{Motivations and High Level Considerations}

Our primary objective is to develop a deep learning model capable of reliable ECG authentication. Instead of using off-the-shelf networks and transfer learning, we analyzed the specific challenges that come with this problem and designed our model accordingly.

\subsection{Compact Model with Broad Receptive Field}
The feasibility of deploying an ECG biometric system depends on the model being lightweight. This requirement is imposed for several reasons. Firstly, the end goal would be to run the model on wearable devices, as smartwatches now carry ECG sensors. Since the typical wearable processors can not be made powerful due to space and power constraints, making the model simple is crucial. Secondly, when new users register to the system, they are likely to provide only a few ECG samples, therefore it would be difficult to obtain a reasonable amount of samples to properly fine-tune a deep network.

Although it is motivating and suitable to design a shallow network, such networks are likely to fall behind in performance. Deep networks have a broader field of vision at the later layers and thus they can make much complex inferences. This ability to analyze a larger portion of the complex featuremaps comes at the cost of network complexity. However, in recent times a few approaches have been proposed to mitigate this. MultiRes blocks \cite{ibtehaz2020multiresunet} can be used interchangeably with convolutional layers, providing a wider receptive field through multiresolution analysis, in a compact form. In terms of pooling, Spatial Pyramid Pooling (SPP) \cite{he2015spatial} can be used to leverage pooling operations from different points of view, as if they were computed at much deeper layers.

Therefore, we use MultiRes blocks and SPP layers to design a compact network with the essence of a broad field of view found in deeper networks. More details are presented in Section \ref{sec:net_arch}.

\subsection{Adapting Siamese Network Distance}

In order to solve the problem of `Identity Verification', the standard pipeline is to extract features or embeddings using a deep learning model and compute the distance with the templates, based on thresholding which the verification is performed. The distances are computed using either Euclidean distance (L2 distance) \cite{chu2019ecg} or cosine similarity \cite{labati2019deep}. 

\begin{figure}[h]
     \centering
     \begin{subfigure}[b]{0.31\textwidth}
         \centering
         \includegraphics[width=\textwidth]{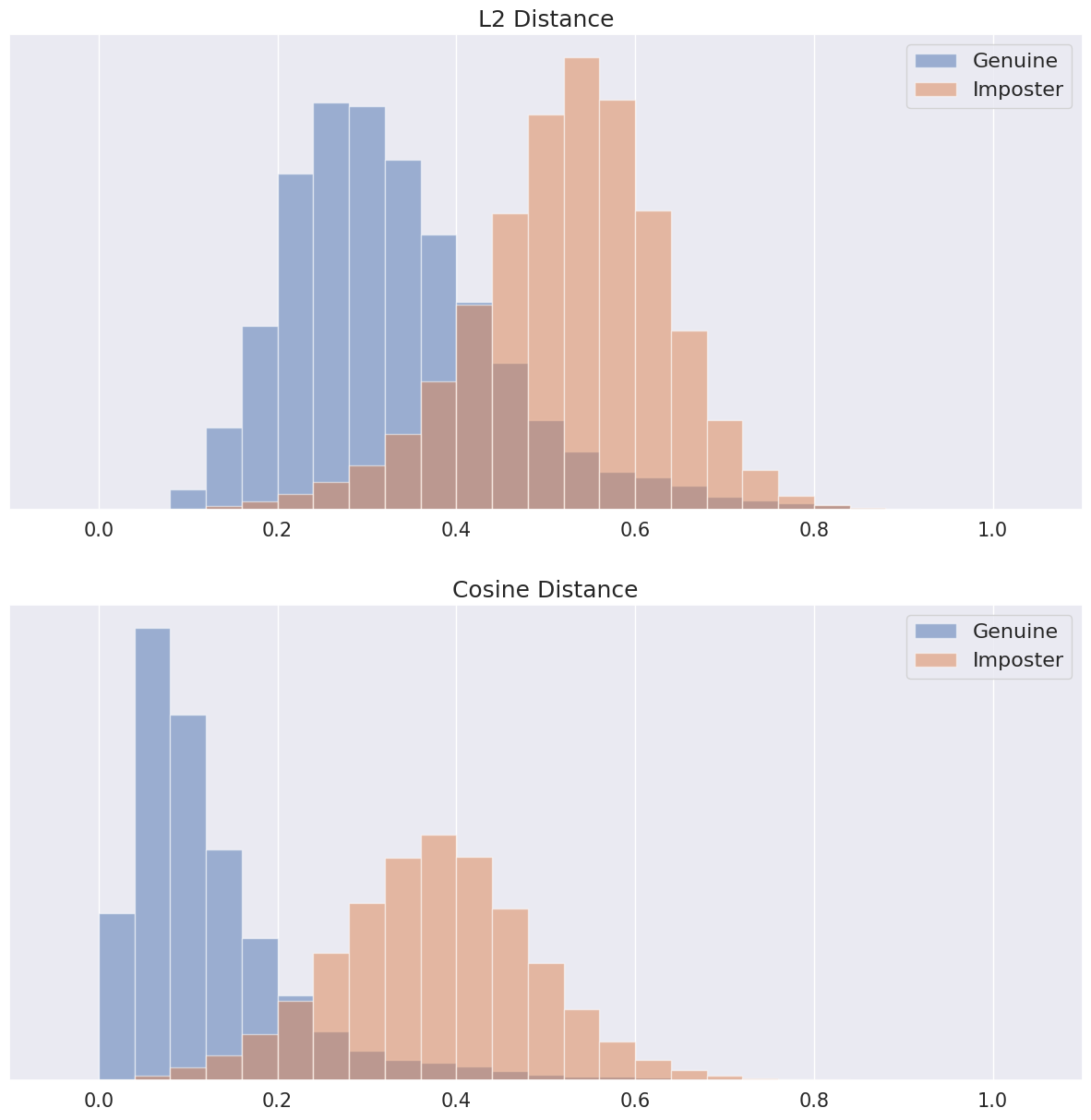}
         \caption{}
         \label{fig:siameseas}
     \end{subfigure}
     \begin{subfigure}[b]{0.31\textwidth}
         \centering
         \includegraphics[width=\textwidth]{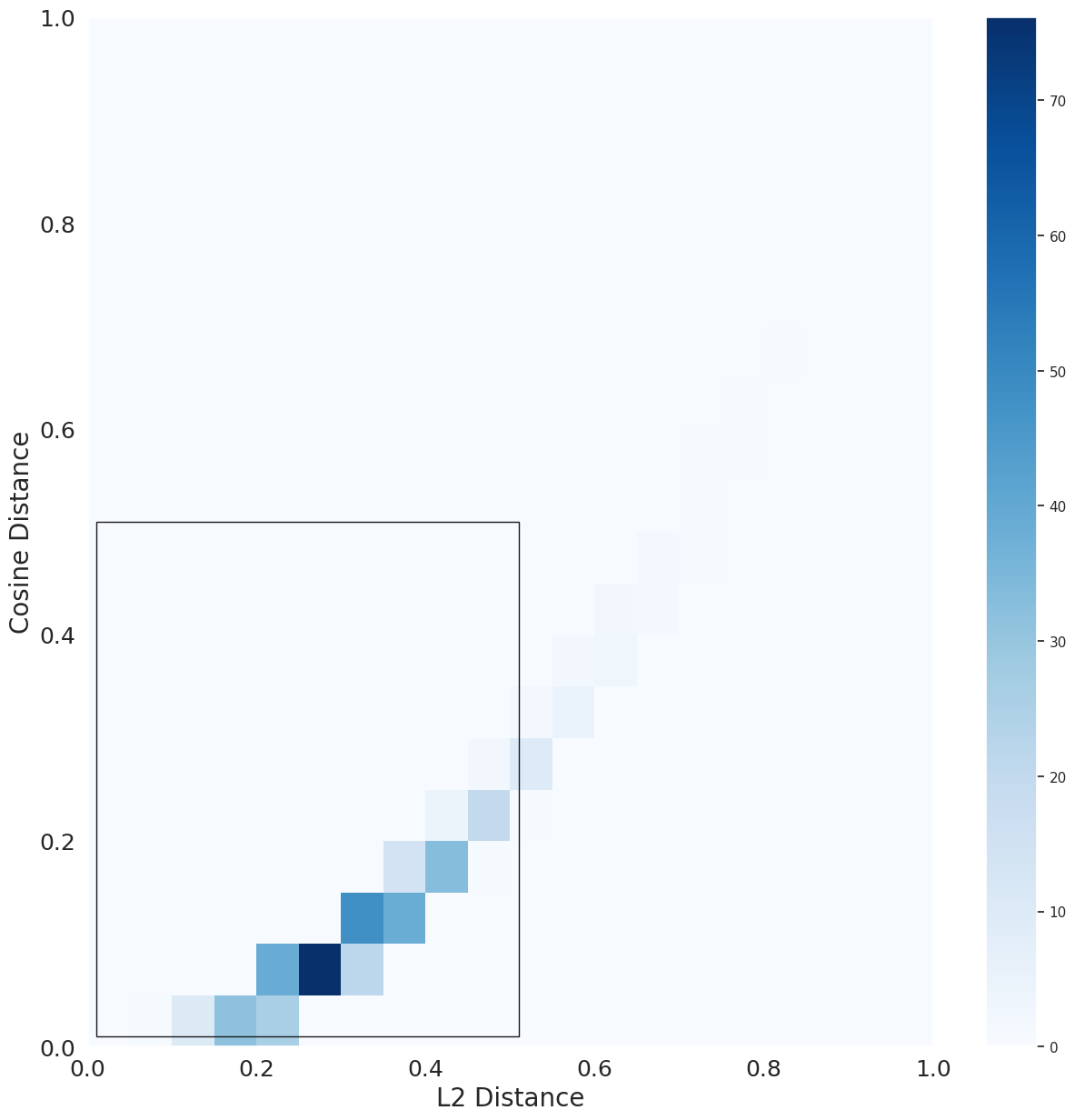}
         \caption{}
         \label{fig:siamese_genas}
     \end{subfigure}
     \begin{subfigure}[b]{0.31\textwidth}
         \centering
         \includegraphics[width=\textwidth]{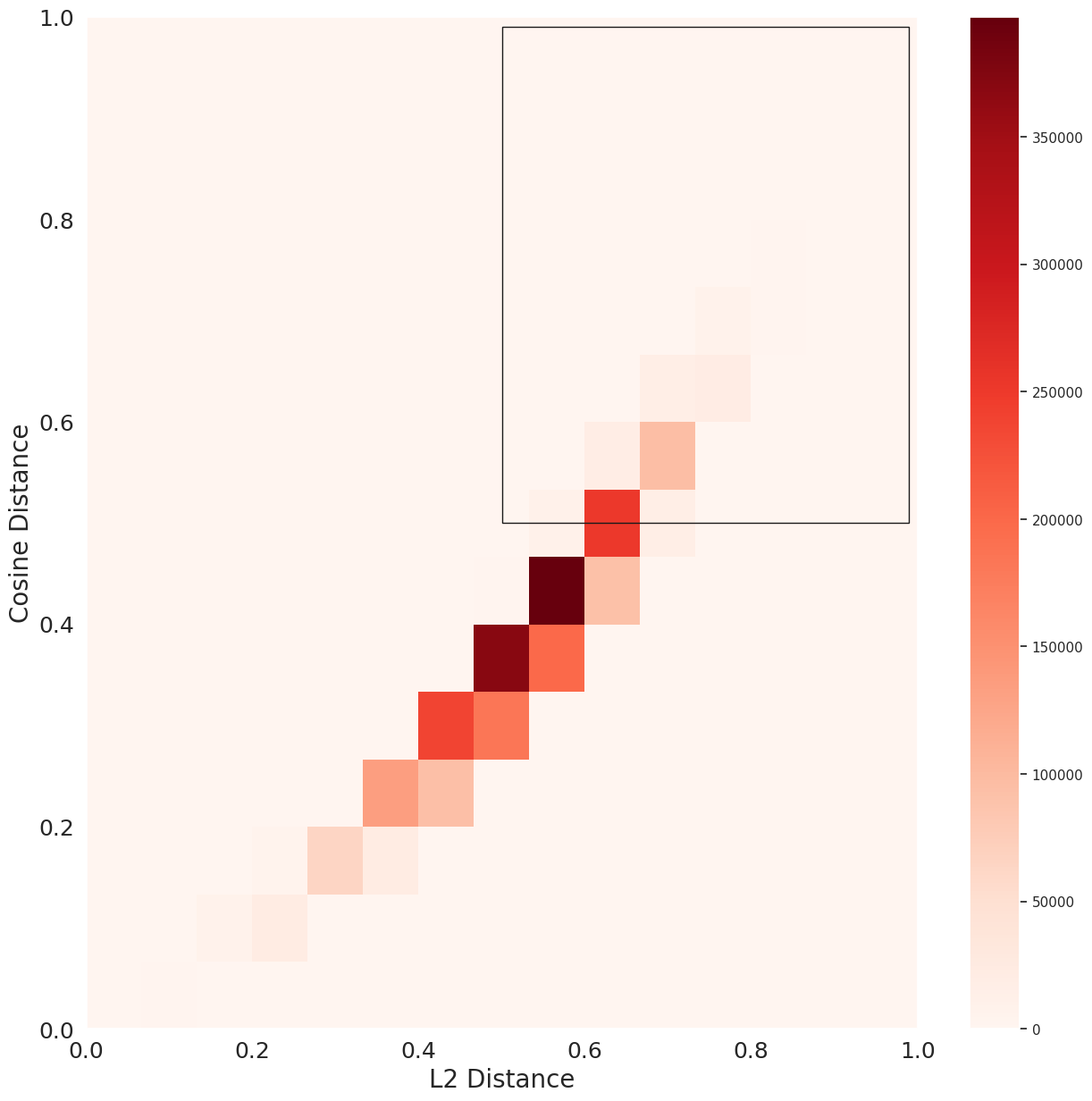}
         \caption{}
         \label{fig:siamese_impas}
     \end{subfigure}
        \caption{Investigation of Distance Measures.}
        \label{fig:eer}
\end{figure}

Cosine similarity has been the defacto standard in identity verification. To investigate the difference in applicability between the two, we extract embeddings from ECG-ID dataset signals and measure both the cosine distance $(1- cosine\,\, similarity)$ and L2 or Eucledian distance, and plot histograms (Fig. \ref{fig:siameseas}). It is evident that L2 distance has comparatively more overlaps, thus the use of cosine similarity is justified. However, when we plot both the distances in a 2D histogram, separately for genuine (Fig. \ref{fig:siamese_genas}) and imposter (Fig. \ref{fig:siamese_impas}) comparisons, we see a different view. In both the figures, the expected region is pointed by a black border and it can be observed that cosine distance is better at distinguishing the genuine ones whereas L2 distance is more suitable for identifying imposters. 

Therefore, we design a novel Siamese network modeling both the two types of distance. We hypothesize that an adaptive combination of both the distances, learned by the Siamese network will be capable of verifying identities better. More details have been presented in Section \ref{sec:siam_arch}.

\subsection{Multiple Beat Fusion}

Although we would like to authenticate the user using a single heartbeat signal, just like a single fingerprint reading can do, it fails to achieve satisfactory performance. Not only a single heartbeat signal contains very little information, but also it succumbs to deviation due to several variabilities. As a result, in the literature, the common practice is to work with multiple beats and it greatly boosts the authentication accuracy \cite{salloum2017ecg,ihsanto2020fast,chu2019ecg}.

When it comes to working with multiple beats, two approaches are followed in general. One approach is to merge multiple beats together and train the model with longer signals \cite{salloum2017ecg,chu2019ecg}. The second approach is to train the model using single beats but during inference, multiple consecutive beats are taken as input and a majority voting of their predictions is provided as the final output \cite{ihsanto2020fast}.

In our work, we followed the second approach, i.e., we train the model with single heartbeats but during inference, we have the option to ensemble results from multiple consecutive heartbeats using majority voting. Our rationale is based on a few considerations. Firstly, training the model on longer signals would require the model to be more complex but we are less likely to have sufficient data to train such a model properly. Secondly, not only it is known that ensemble methods always boost the accuracy, but also this provides us with the flexibility to work with a single beat if needed (e.g. continuous authentication \cite{du2015continuous}). And most importantly, if even one beat deviates from the standard morphology due to some physiological reasons, it will negatively affect the entire merged multibeat signal (i.e., approach 1), but ensembling predictions from multiple beats will be comparatively immune to such inconsistencies. This is our primary motivation, to make the method more robust.

\section{Methodology}

\subsection{Problem Definition}
The problem of individual authentication using ECG signals can be divided into two tasks: Closed Environment Identification and Identity Verification. 

\subsubsection{Closed Environment Identification}
For the closed environment identification problem, among a fixed set of $n$ persons, $P=\{p_1,p_2,p_3,\dots,p_n\}$. The task is to develop a model $M_{cei}$ that is capable of identifying the person $p_i$ when given an ECG signal of that person $Sig_i$. This task, eventually reduces into a multi-class classification problem.

\subsubsection{Identity Verification}
\label{sec:siam_arch}

One major drawback of the former task is that when a new individual is introduced to the authentication system, the entire pipeline needs to be retrained with the new data. This becomes cumbersome for practical usage. Therefore, an alternate task has also been studied in the literature. In the identity verification task, we are given an independent dataset of persons ($P_i ={p_{i1},p_{i2},p_{i3},\dots,p_{im}}$) and based on a certain portion of data, termed as `enrollment data' we generate templates ($T ={t_{t1},t_{2},t_{3},\dots,t_{m}}$) for them. Then, the task is to match the remaining data, termed as `evaluation data' with the templates and verify the identity. To match a sample $Sig_a$ with the template $t_b$, i.e. measuring the similarity or dissimilarity, distances like cosine %($1-\frac{Sig_a \cdot t_b}{||Sig_a|| \; ||t_b||}$ 
or Euclidean distances %($\sqrt{\sum_{p=1}^{q}()}$) 
 are used.

%we are given two disjoint set of persons, $P_a=\{p_{a1},p_{a2},p_{a3},\dots,p_{am}\}$ and $P_b=\{p_{b1},p_{b2},p_{b3},\dots,p_{bm}\}$ . A model $M_{iv}$ is trained using the data from $P_a$ and then tested on unseen $P_b$. For this task, instead of performing any classification, the purpose of the model is to generate embeddings. The data from $P_b$  is further divided into two partitions, one being the enroll data $P_{{enroll}_b}$ and the remaining being the evaluation data $P_{{eval}_b}$. The enroll data is used to generate templates for the individuals from $P_b$

\subsection{Datasets}

In order to develop, experiment, and evaluate the EDITH system, we consider 4 different public datasets, which are the main ECG datasets used in ECG based authentication purposes in the literature \cite{pinto2018evolution}. Below we briefly describe these datasets (please see Table \ref{tbl:data} for an overview). 

\subsubsection{ECG-ID}

The ECG-ID database \cite{lugovaya2005biometric,nemirko2005biometric} is one of the few databases that was developed for the sole purpose of investigating ECG-based authentication. Therefore, this dataset is the defacto choice for researchers to work with ECG biometrics \cite{pinto2018evolution}. In this dataset, 20 seconds long ECG signal from lead I, was recorded from 90 persons. The signals were digitized at 500 Hz with 12-bit resolution over a nominal $\pm10$ mV range. For each subject, 2 to 20 session data were collected. Thus it is the common practice among researchers to consider only 2 sessions for each person to avoid any data imbalance issue \cite{ihsanto2020fast,salloum2017ecg}, which is also considered in this study.

\subsubsection{MIT-BIH Arrhythmia}
Although the MIT-BIH Arrhythmia dataset \cite{moody2001impact} is centered around arrhythmia detection, this dataset has been used to benchmark biometric authentication accuracy in several works. This dataset contains ECG signals from 47 subjects suffering from a wide variety of arrhythmias. The recordings were digitized at 360 Hz sampling rate with 11-bit resolution over $10$ mV range of signals. The ECG signals were recorded in two channels, however, only the first channel was used for this study.

\subsubsection{PTB Diagnostic ECG Database}

The PTB Diagnostic ECG database \cite{bousseljot1995nutzung} contains 549 records from 290 subjects, out of which 52 individuals are healthy. Thus, for benchmarking authentication performances only this control group has been used in the literature \cite{chu2019ecg}. This dataset contains ECG records from 15 leads (12 conventional leads and 3 Frank leads). The signals are digitized in 1000 Hz sampling frequency with 16 bits resolution. The ECG signals from the Frank lead (vx) was used in this study.

\subsubsection{MIT-BIH NSRDB}

The NSRDB database is a part of the MIT-BIH Arrhythmia database \cite{moody2001impact}. This database contains signals from 18 subjects, who are free from arrhythmias or other abnormalities. This database mostly contains Normal Sinus Rhythm, thus the name NSRDB. 1000 beats from each of the 18 individuals were considered for this study.

All the datasets were collected from PhysioNet \cite{goldberger2000physiobank} and were resampled to 500 Hz (in case of a different sampling rate). Notebly all these datasets were already filtered, thus no additional filtering or preprocessing was applied. However, when applying the proposed method to a different data source, some preprocessing may be necessary to remove noise and artifacts.

An overview of the datasets has been presented in Table \ref{tbl:data}

\begin{table}[h]
\caption{Overview of the Datasets}
\label{tbl:data}
\footnotesize
\begin{tabular}{|l|c|c|c|c|c|l|}
\hline
Dataset & \# Persons & \begin{tabular}[c]{@{}c@{}}Sampling\\ Rate\end{tabular} & \begin{tabular}[c]{@{}c@{}}Health\\ Condition\end{tabular} & Activity & \begin{tabular}[c]{@{}c@{}}Electrode\\ Placement\end{tabular} & \multicolumn{1}{c|}{Reference} \\ \hline
ECG-ID & 90 & 500 Hz & Healthy & Sitting & Wrist & \cite{lugovaya2005biometric,nemirko2005biometric} \\ \hline
\begin{tabular}[c]{@{}l@{}}MIT-BIH\\ Arrhythmia\end{tabular} & 47 & 360 Hz & Arrhythmia & \begin{tabular}[c]{@{}c@{}}Ambulatory\\ Recording\end{tabular} & Chest & \cite{moody2001impact} \\ \hline
PTB & 290 & 100 Hz & Mixed & At rest & Chest + Limbs & \cite{chu2019ecg} \\ \hline
\begin{tabular}[c]{@{}l@{}}MIT-BIH\\ NSRDB\end{tabular} & 18 & 360 Hz & Healthy & \begin{tabular}[c]{@{}c@{}}Ambulatory\\ Recording\end{tabular} & Chest & \cite{moody2001impact} \\ \hline
\end{tabular}
\end{table}

In order to perform cross-session evaluation, we used another dataset, CYBHi \cite{da2014check}.

\subsubsection{CYBHi Dataset}

The CYBHi \cite{da2014check} dataset was developed to evaluate ECG biometrics, particularly in multi-session scenario. The dataset contains two types of data, short-term and long-term. In the long-term dataset, ECG signals from 63 individuals are presented which are collected over a period of several days. The ECG signals were collected from finger using a sampling frequency of 1 kHz.

\subsection{Proposed Method}

\begin{figure}[h]
    \centering
    \includegraphics[width=\textwidth]{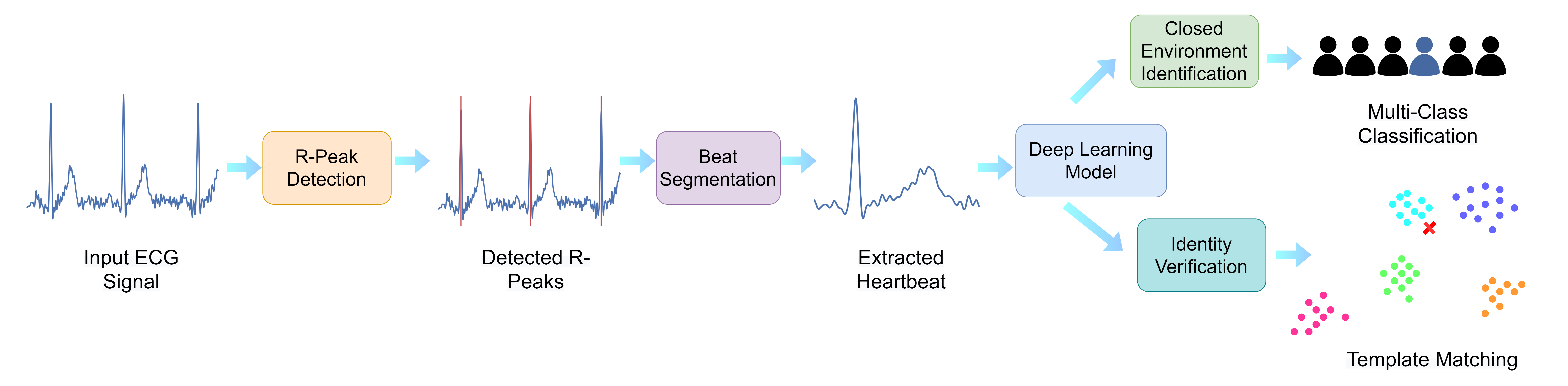}
    \caption{The proposed EDITH Pipeline. From an input ECG signal the R-peaks are detected using deep learning (MultiResUNet) model, based on which the heartbeats are segmented. The extracted beats are analyzed using another deep learning (CNN or Siamese) model for closed environment identification or identity verification.}
    \label{fig:pipeline}
\end{figure}

The proposed EDITH pipeline takes a continuous ECG signal as input ($f_s = 500 Hz$, resamples otherwise) and detects the R-peaks, based on which the beats are extracted. Each heartbeats is then passed through a convolutional neural network (CNN) when dealing with the closed environment identification task. On the contrary, to perform the identity verification task they are compared with the stored templates using a siamese network. Furthermore, the results from multiple beats can be ensembled for improving the performance. An overview of the pipeline is presented in Fig. \ref{fig:pipeline}.

\subsubsection{R-peak Detection}
The first step of the proposed method is the R peak detection. It is often attributed that R peaks being the highest and sharpest point are trivial to detect \cite{chu2019ecg}. However, for a diverse set of ECG signals, this is not always the case. As has been demonstrated by recent works \cite{ihsanto2020fast}, popular beat detection algorithms (e.g. Hamilton \cite{hamilton2002open}) are not consistently sufficient to precisely locate the R peaks. It was also demonstrated that when working with a single-beat authentication system, proper detection of R-peaks is the most vital step, deviating even slightly often diminishes the performance drastically \cite{ihsanto2020fast}.

Uzair \textit{et al.} \cite{uzair2020robust} have recently proposed an approach to detect R-peaks using deep learning technique, which outperforms the traditional methods. In that work, a 1-D U-Net \cite{ronneberger2015u} was used to predict the location of R-peaks, from an ECG signal. We have deployed our recently proposed variant of U-net architecture, 1-D MultiResUNet model \cite{ibtehaz2020multiresunet} with deep supervision \cite{lee2015deeply} as done in \cite{ibtehaz2020ppg2abp}, to predict the R-peaks from a continuous ECG signal. MultiResUNet model showed improved performance for 2D applications over the traditional U-Net architecture which incorporates multiresolution analysis and deep supervision is proven to improve the performance of deep networks. The probability map predicted from the model is thresholded and further refined by some post-processing. It is expected that R-peaks will be considerably distant from one another. Thus, to minimize false positives, we ignore nearby predicted beats and replace them with the median prediction. To reduce false negatives we empirically select a suitable threshold value.

\subsubsection{Beat Segmentation}

After the R-peaks have been detected, the individual heartbeats are segmented from the ECG signal, following the approach in \cite{ihsanto2020fast}. The beats are aligned based on the location of the R-peaks to provide a standardized pattern for the model. We take a window size of $w$, where $w=256$ samples, and include $\frac{w}{4}$ and $\frac{3w}{4}$ samples before and after the R-peak respectively. Furthermore, the extracted heartbeat signals were Z-score normalized \cite{ihsanto2020fast}.

\subsubsection{Network Architecture}
\label{sec:net_arch}
 
Developing a deep learning model for a single heartbeat based authentication system is challenging and difficult for two major reasons: short signal duration and the requirement of a large database. Short signals will be affected by pooling layers, for losing probable significant information. Furthermore, a small(er) dataset prevents us from working with deeper networks. Consequently, pooling operations have been avoided in the literature for this purpose \cite{ihsanto2020fast} and multiresolution analysis has rather been leveraged instead \cite{zhang2017heartid,chu2019ecg}.

\begin{figure}[h]
     \centering
     \begin{subfigure}[b]{0.43\textwidth}
         \centering
         \includegraphics[width=\textwidth]{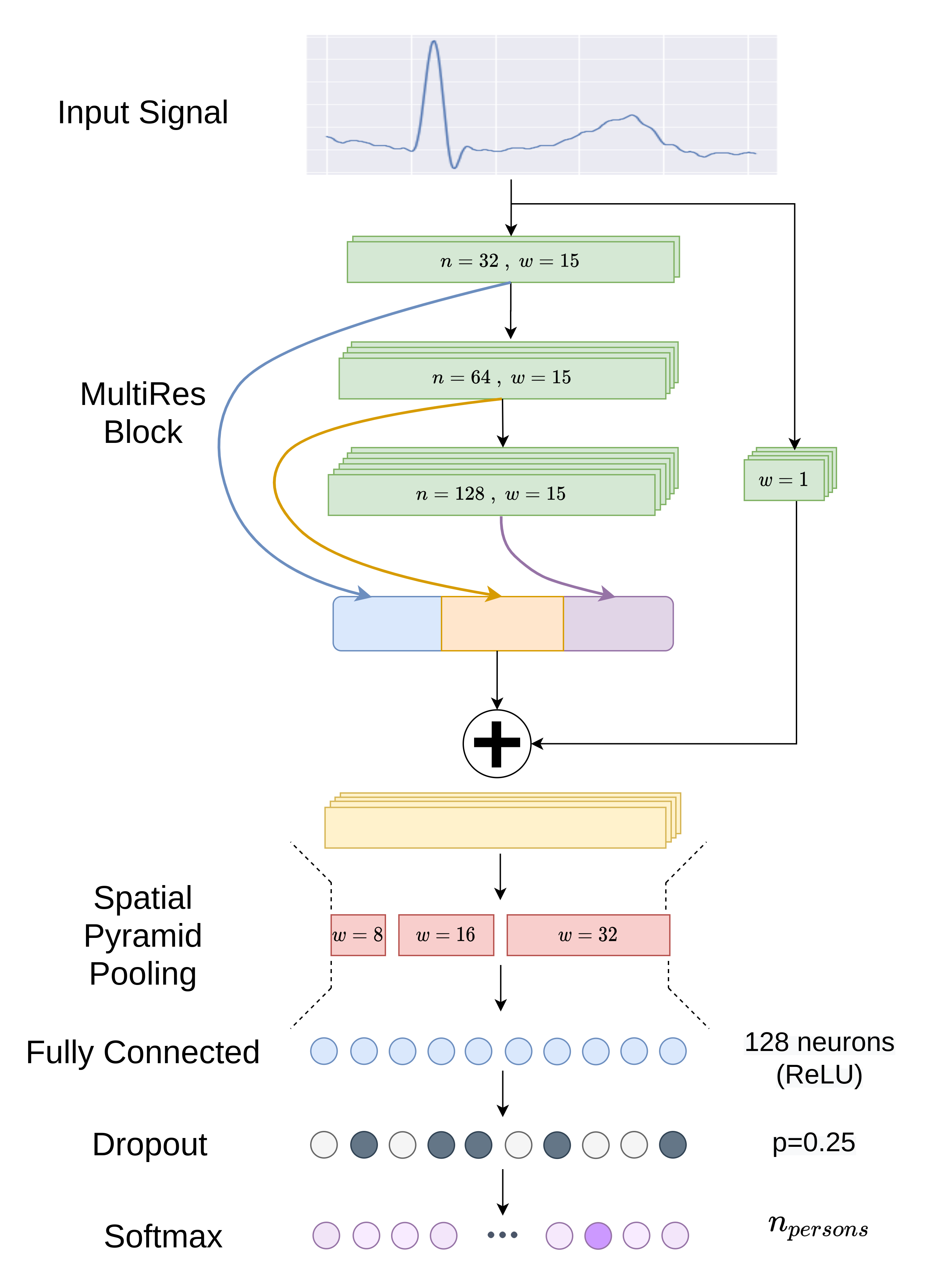}
         \caption{}
         \label{fig:net_archi1}
     \end{subfigure}
     \hfill
     \begin{subfigure}[b]{0.56\textwidth}
         \centering
         \includegraphics[width=\textwidth]{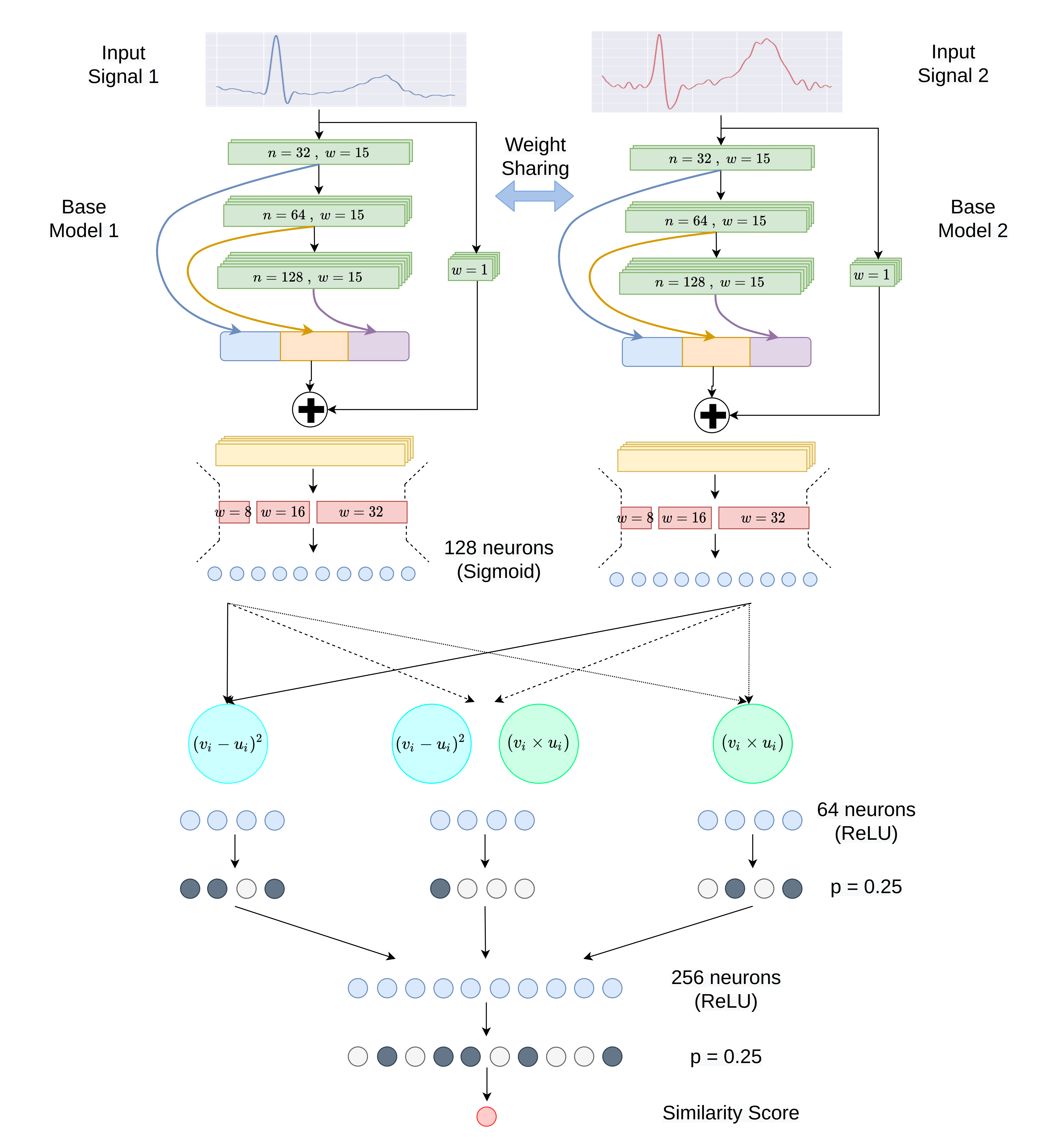}
         \caption{}
         \label{fig:net_archi2}
     \end{subfigure}
        \caption{\textbf{The proposed network architecture.} \ref{fig:net_archi1} presents the Closed Environment Identification model and \ref{fig:net_archi2} refers to the proposed Siamese model for Identity Verification.}
        \label{fig:net_archi}
\end{figure}

Therefore, in our proposed model, we have resorted to using a somewhat shallow network, incorporating multiresolution analysis with limited pooling operations. We were motivated to use the MultiRes block as proposed in \cite{ibtehaz2020multiresunet} and spatial pyramid pooling \cite{he2015spatial}. The network architecture has been presented in Fig. \ref{fig:net_archi1}.

The MultiRes block presents a compact form of Inception block \cite{szegedy2015going} and incorporates a residual connection for aiding in the gradient flow \cite{he2016deep}. Both inception block \cite{chu2019ecg} and residual connections \cite{ihsanto2020fast} have demonstrated favorable outcome in ECG based biometrics. We place 3 convolutional layers having filters of size 32, 64, and 128, respectively and concatenate their featuremaps, which in essence factorizes larger convolutional operations using smaller ones. Since we are using a shallow network, we consider convolutional filters of size 15, to expand the receptive field. We avoid intermediate pooling operations to prevent loss of information, but after performing the convolutions for the sake of summarization of the featuremaps, max pooling is performed. To exploit the multiresolutional nature of our analysis further, we perform spatial pyramid pooling instead of the vanilla pooling operation. Three pooling operations, with windows of size of 8, 16, and 32 are performed in parallel and the outputs are concatenated. Finally, a fully connected layer with 128 neurons is introduced to generate the embedding for the input signals. This layer is further regularized with a dropout of 0.25 to avoid overfitting \cite{srivastava2014dropout}. All the layers used in this network are batch-normalized \cite{ioffe2015batch} and activated by Rectified Linear Unit (ReLU) activation.

\subsubsection{Closed Environment Identification}

For the Closed Environment Identification task, the target is to identify a predefined number of individuals. Therefore, the problem reduces to a Multi-Class Classification problem. We simply include a softmax layer on top of our proposed network to perform this (Fig. \ref{fig:net_archi1}). The number of neurons at the output layer is kept equal to $n_{persons}$, i.e. the total number of subjects in the closed environment being considered.

\subsubsection{Identity Verification}

For the identity verification task, on the other hand, we do not have any fixed number of individuals as our targets; rather a set of templates are computed to compare with. In the literature, for this task, typically a group of subjects is used to train a deep learning model, which is then used to generate embeddings, by removing the top layers. Then, a set of enrollment signals is considered, containing a few signals from each of the individuals to evaluate the performance of the method. Templates are computed for the individuals by taking an average of the embeddings of that individual.

Distance measures like Euclidean distance \cite{chu2019ecg} or cosine similarity \cite{labati2019deep} are widely used in the literature, to match test signals with the templates. We hypothesize that such distance measures may not be capable of weighting the individual features of the embedding properly, and thus an adaptive method may be more suitable.

To this end, we propose a Siamese architecture to compute the similarity between templates and signal embeddings. Siamese networks employ two identical sub-networks to compare the similarity of two input samples, and they have a long history in performing verification tasks \cite{bromley1993signature}. The weights of the two sub-networks are shared together, and hence, they are likely to predict similar featuremaps for similar input samples. They can be further compared together, usually by computing the squared differences between the individual pairs of features. However, drawing motivation to emulate the popular Cosine similarity distance metric used for this purpose, in addition to computing the squared differences, we also explore the efficacy of multiplying the individual pairs of features. We term this as \textit{Product Proximity} between the features, as similar valued feature pairs will yield a high multiplication output, whereas feature pairs with dissimilar values will result in a low score. Our intuition is that the way squared difference is intended to model eucledian distance, \textit{Product Proximity} of the features will represent correlation or Cosine similarity like metrics. In addition to working with them separately, we also merge them together and analyze them in fully connected layers (please refer to Fig. \ref{fig:net_archi2}). Mathematically,

\begin{equation}
    \textit{Squared Difference}[i] = (v_i - u_i)^2, \forall i \in [0\dots d]
\end{equation}
\begin{equation}
    \textit{Product Proximity}[i] = (v_i \times u_i), \forall i \in [0\dots d]
\end{equation}
\begin{equation}
    \textit{Combined Metric}[i] = \textbf{concatenate}(\textit{Squared Difference}[i],\textit{Product Proximity}[i])
\end{equation}

Here, $v$ and $u$ are $d$-dimensional embeddings generated from a model $\phi$ for inputs $U$ and $V$ respectively, i.e., $v = \phi(V)$, $u = \phi(U)$.

Hence, we take our proposed identification network, pretrained on a disjoint set of individuals. The top softmax layer and the dropout layers are then discarded and the top ReLU activation is replaced with a Sigmoid activation instead so that we can obtain the embeddings in the range of $0\sim1$. We take another identical sub-network so that the embeddings for two signals can be computed in parallel. After the embeddings are computed, we apply two different types of distance measures between the individual features as mentioned earlier and we further perform a combination of the two. The distance measures are further analyzed by the subsequent fully connected layers and by merging them we predict a similarity score between $0 \sim 1$. The model architecture for the identity verification task has been presented in Fig. \ref{fig:net_archi2}.

\section{Experimental Setup}

The experiments have been conducted in a server computer with Intel Xeon @2.2GHz CPU, 24 GB RAM, and NVIDIA TESLA P100 (16 GB) GPU. We implemented the EDITH pipeline in Python programming language. The deep learning models were implemented in Keras \cite{chollet2015keras} with Tensorflow backend \cite{abadi2016tensorflow}. The codes are available in the following github repository.

\begin{center}
    \url{https://github.com/nibtehaz/EDITH}
\end{center}

\subsection{R-peak Detection Model}

In order to train the R-peak Detection Model, we collected the portion of the ECG signals with ground truth annotation of the R-peaks from the ECG-ID  database. The continuous signals were divided into windows of 1024 samples, with a step size of 256 samples. Each windowed signal is composed of roughly three ECG beats. These windowed signals were considered as input to the R-peak detector model, while the output was a pulse train showing the location of R-peaks as `1' and `0' elsewhere. In order to apply deep supervision, as reported in \cite{ibtehaz2020ppg2abp}, the pulse trains were subsampled accordingly. 80\% of the data was used to train the model and the remaining 20\% was used as validation data. The model was trained for 100 epochs, minimizing binary cross-entropy loss using the Adam optimizer \cite{kingma2014adam}.

\subsection{Closed Environment Identification Model}
\label{sec:cei}
For the Closed Environment Identification model, we performed stratified 10-fold cross-validation as well as testing using 60-20-20 train-val-test split. Thus the data were divided and the CNN models were trained accordingly. We trained the models for 500 epochs, minimizing categorical crossentropy loss using Adam optimizer \cite{kingma2014adam}. Notably, except for the PTB dataset, convergence were reached before 150 epochs for all. 

The models were evaluated using the standard accuracy metric as follows:

\begin{equation}
    Accuracy = \frac{TP+TN}{TP+TN+FP+FN}
\end{equation}

Here, TP, TN, FP, FN corresponds to the true positive, true negative, false positive, false negative predictions respectively.

In addition, we also compute other standard evaluation metrics:

\begin{equation}
    Precision = \frac{TP}{TP+FP}
\end{equation}
\begin{equation}
    Recall = \frac{TP}{TP+FN}
\end{equation}
\begin{equation}
    F1-Score = \frac{2}{\frac{1}{Precision}+\frac{1}{Recall}}
\end{equation}

\subsection{Identity Verification Model}

We trained the base model (as mentioned in Section \ref{sec:cei}) using 50\% of the individuals from the ECG-ID dataset. 

The training signals were paired with each other to form training samples for the Siamese model. As we have too many mismatched samoles in comparison to matched samples, we used Synthetic Minority Oversampling Technique (SMOTE) \cite{chawla2002smote} to generate synthetic data for matching cases. 21.5 times synthetic samples were generated for each person as a trade-off between data imbalance and computation requirements. This enables the model to learn which patterns are discriminative of matching cases. 80\% of the signal pairs were used to train the model and the rest were used as validation data. Since we predict a continuous variable as the outcome in this model the Adam optimizer \cite{kingma2014adam} was used to minimize the Mean squared error (MSE) loss. The model was trained for 75 epochs to predict `0' for mismatches and `1' for matches.

For evaluation, the test data was later split into enrollment and evaluation data. The enroll signals were used to generate embeddings, and for each subject, the mean or centroid of the embeddings was considered as the template. The evaluation signals were matched with the templates for each individual, and based on a threshold the match-mismatch decision was made. Correspondingly the standard evaluation metrics, namely False Acceptance Rate (FAR), False Rejection Rate (FRR), and Equal Error Rate (EER) were computed.

\begin{equation}
    FAR(th) = \frac{False Acceptance(th)}{n_{Imposter}}
\end{equation}
\begin{equation}
    FRR(th) = \frac{False Rejection(th)}{n_{Genuine}}
\end{equation}
\begin{equation}
    EER = FAR(th_{equal}) 
\end{equation}
    $$\exists th_{equal} \in [0\dots1] \land FAR(th_{equal})=FRR(th_{equal})$$

The threshold value $th$ were taken from 0 to 1, and the threshold  $th=th_{equal}$ causing $FAR(th)=FRR(th)$, was used to compute $EER$.

\section{Results and Discussions}

%\subsection{EDITH reliably detects R-peaks}
\subsection{R-peak detection by EDITH}

Although detection of R-peaks has not been our primary focus, it turns out that our proposed R-peak detection model detects R-peaks with reliable performance. Despite being an auxiliary step of our pipeline, it is on par with the popular R-peak detection algorithms \cite{hamilton2002open,christov2004real,engelse1979single,pan1985real,kalidas2017real,elgendi2010frequency}, and outperforms most both quantitatively and qualitatively.

To present a comparison with the existing algorithms, we detect the R-peaks on the entire ECG-ID and benchmark the various algorithms based on the ground truth annotations provided therein. The results are presented in Table \ref{tbl:rpeak}. We consider the false positives and false negatives as falsely predicted peaks and actual peaks that were missed respectively. Besides, to further assess the time lag of R-peak detection, we propose a metric termed `Temporal Error'. Suppose, a certain R-peak, $r_i$ was at timestamp $t_i$ and the method predicts it at timestamp $\hat{t}_i$, the Temporal Error is defined as, $Temporal\;Error = |\hat{t}_i-t_i|$. The proposed R-peak detector achieves the least number of false positives, and quite remarkably a negligible Temporal Error. This ensures a proper segmentation of the heartbeats for authentication in the next stage of our pipeline.

\begin{table}[h]
\caption{A quantitative evaluation of the proposed R-peak detector. EDITH predicts the least number of false negatives and maintains a negligible Temporal Error.}
\label{tbl:rpeak}
\scriptsize
\begin{tabular}{|
>{\columncolor[HTML]{C0C0C0}}lV{4}c|c|c|c|c|c|c|}
\hline
Method & Proposed & Hamilton \cite{hamilton2002open} & Christov \cite{christov2004real} & \begin{tabular}[c]{@{}c@{}}Engelse-\\ Zeelenberg \\ \cite{engelse1979single}\end{tabular} & \begin{tabular}[c]{@{}c@{}}Pan-\\ Tompkins \\ \cite{pan1985real}\end{tabular} & \begin{tabular}[c]{@{}c@{}}Stationary \\ Wavelet \\ Transform \cite{kalidas2017real}\end{tabular} & \begin{tabular}[c]{@{}c@{}}Two \\ Moving \\ Average\cite{elgendi2010frequency}\end{tabular} \\ \hline
Detected R-peaks & 2904 & 2873 & 2898 & 2661 & 2883 & 2887 & 2903 \\ \hline
False Positives & 132 & 197 & 144 & 109 & 183 & 168 & 173 \\ \hline
False Negatives & 196 & 227 & 202 & 439 & 217 & 213 & 197 \\ \hline
Temporal Error & $0.69 \pm 5.87$ & $37.86 \pm 39.41$ & $12.38 \pm 24.69$ & $3.82 \pm 20.07$ & $25.61 \pm 33.57$ & $27.64 \pm 32.44$ & $30.28 \pm 22.07$ \\ \hline
\end{tabular}
\end{table}

\begin{figure}[h]
     \centering
     \begin{subfigure}[b]{\textwidth}
         \centering
         \includegraphics[width=\textwidth]{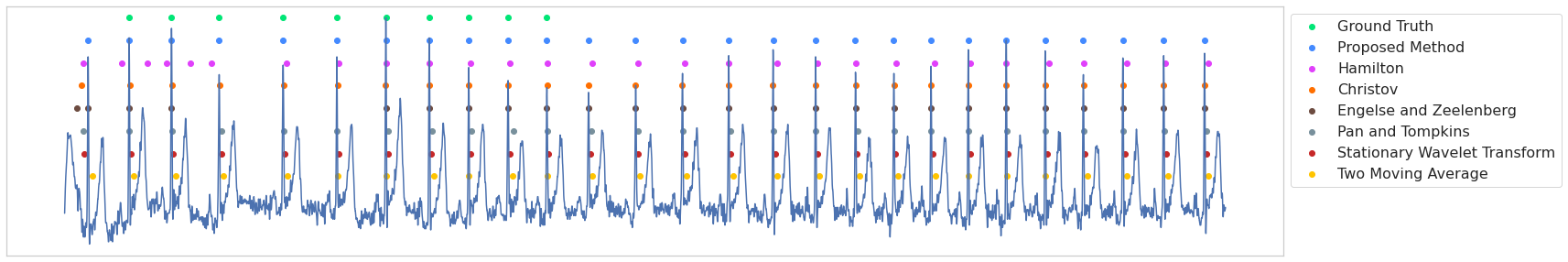}
         \caption{}
         \label{fig:r_res}
     \end{subfigure}
     
     \begin{subfigure}[b]{\textwidth}
         \centering
         \includegraphics[width=\textwidth]{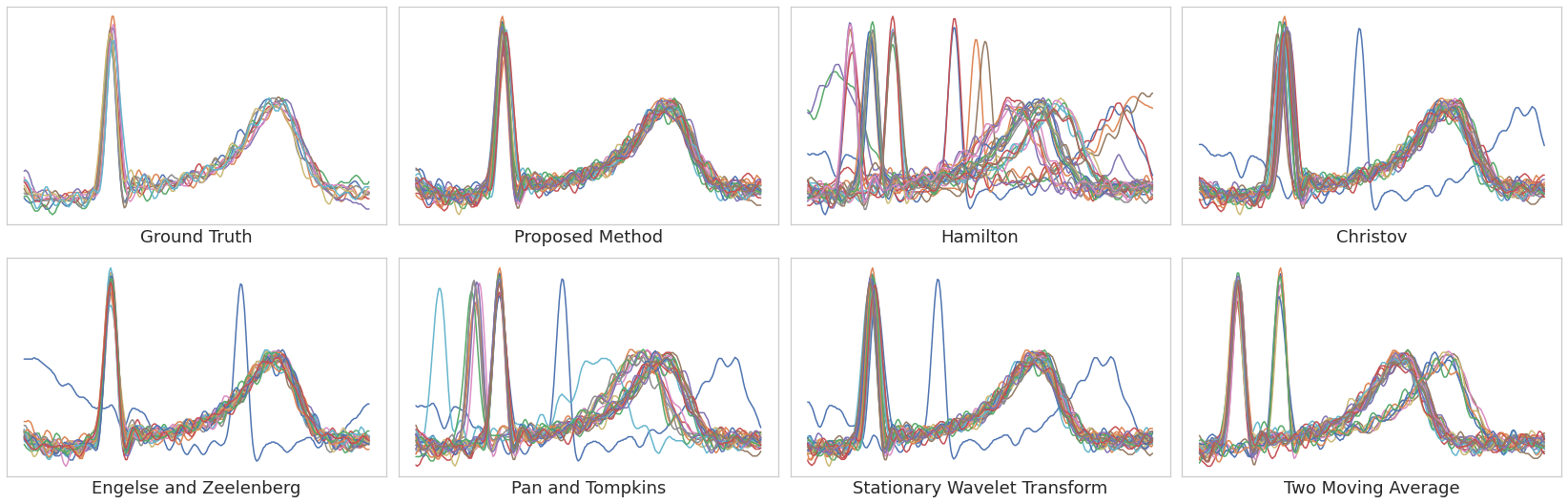}
         \caption{}
         \label{fig:r_segments}
     \end{subfigure}
        \caption{A qualitative evaluation of the proposed R-peak detector. \ref{fig:r_res} presents the R-peaks detected by various methods from a sample ECG signal of the ECG-ID database. \ref{fig:r_segments} demonstrates the segmented beats based on the detected R-peaks using those methods.}
        \label{fig:qudata}
\end{figure}

In terms of qualitative assessment of the R-peak detection, we select an ECG signal from the ECG-ID dataset and detect the R-peaks using the different methods. The comparative analysis of the R-peak along with the manual ground-truth annotation (only 10 peaks for each signal is provided in the database) is presented in Fig. \ref{fig:r_res}. Interestingly, the proposed R-peak detector accurately detects all the peaks, whereas other methods predict false peaks or miss R-peaks, and usually yield a high time lag between the detected and the actual peak. The superior performance of the proposed method is further illustrated in Fig. \ref{fig:r_segments}, where similar observations can be drawn over the segmented beats based on the detected R-peaks, which is the primary concern in this context. It is evident that the beats detected by the proposed approach are properly aligned, almost identical to the ground-truth. On the other hand, the traditional methods sometimes not only missed beats but also failed in aligning them properly. Notably, missing beats as well as misaligned ones adversely affects the identification accuracy as pointed out in \cite{ihsanto2020fast}.

Furthermore, we used our R-peak detection model to detect R-peaks from the PTB and NSRDB datasets, without any modification or fine-tuning. In both cases, our model was capable of detecting the R-peaks reliably. However, for the lack of manual ground-truth annotation, it was not possible to evaluate the performance directly. Nevertheless, the high level of accuracy in identifying the individuals implicitly points out to the overall robustness of the R-peak detector.

%\subsection{EDITH demonstrates consistency in closed environment identification}
\subsection{Comparative Evaluations for the Closed Environment Identification}

\subsubsection{Closed Environment Identification}
\begin{figure}[h]
     \centering
     \begin{subfigure}[b]{0.49\textwidth}
         \centering
         \includegraphics[width=\textwidth]{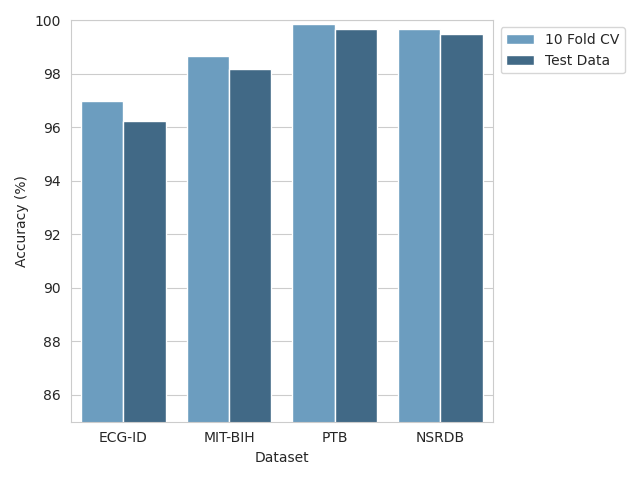}
         \caption{}
         \label{fig:cei_acc}
     \end{subfigure}
     \hfill
     \begin{subfigure}[b]{0.49\textwidth}
         \centering
         \includegraphics[width=\textwidth]{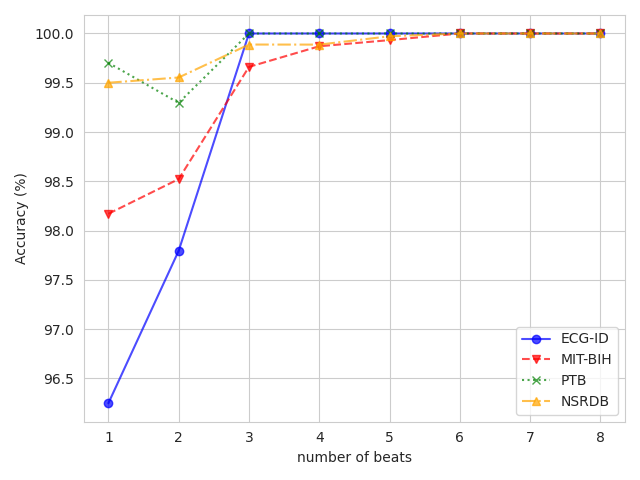}
         \caption{}
         \label{fig:cei_multi}
     \end{subfigure}
        \caption{Closed Environment Identification performance of EDITH. In \ref{fig:cei_acc} the bar plots represent the accuracy obtained using a single heartbeat in each dataset. On the other hand, \ref{fig:cei_multi} illustrates the gain in performance when fusing predictions from multiple beats together.}
        \label{fig:cei}
\end{figure}

In order to assess the efficacy of EDITH for the closed environment identification task, we follow the two traditional evaluation schemes, stratified 10-fold cross-validation, and 60-20-20 data split. EDITH manages to accurately identify the individuals consistently in all datasets. The accuracy plots of EDITH across the four datasets are shown in Fig. \ref{fig:cei_acc}. It can be noticed that for 10-fold cross-validation, the accuracy is slightly higher, which could be due to the availability of additional training data. Nevertheless, in all the experiments, EDITH achieves a remarkable performance in the closed environment identification. In addition, we also present the standard classification metrics in Table \ref{tbl:metrics}. It can be observed that for MIT-BIH, PTB, and NSRDB datasets, the proposed method achieves balanced precision and recall scores which is also reflected at the high F1-Score. Although in the ECG-ID dataset recall is slightly worse, the precision is comparatively higher. This signifies that the method reduces false positives at the cost of false negatives. This proves beneficial as for an authentication system restricting unauthorized access is more concerning. Nevertheless, the recall, i.e. authorized access can be improved by ensembling the result of multiple beats.

\begin{table}[h]
\centering
\caption{Closed Environment Identification train-val-test split single-beat performance}
\label{tbl:metrics}
\begin{tabular}{|c|c|c|c|c|}
\hline
Dataset & Precision & Recall & F1-Score & Accuracy \\ \hline
ECG-ID & 98.053 \% & 94.481 \% & 96.234 \% & 96.247 \% \\ \hline
MIT-BIH & 98.429 \% & 98.072 \% & 98.243 \% & 98.170 \% \\ \hline
PTB & 99.711 \% & 99.702 \% & 99.699 \% & 99.702 \% \\ \hline
NSRDB & 99.586 \% & 99.417 \% & 99.499 \% & 99.500 \% \\ \hline
\end{tabular}
\end{table}

\subsubsection{Fusing Multiple Beats for Improved Performance}
%Although EDITH has indeed achieved impressive accuracy using just a single heartbeat, 
In the literature, it has been demonstrated that combining multiple beats greatly increases authentication accuracy \cite{salloum2017ecg,ihsanto2020fast}. This is expected due to the information gain from the other beats. %As a result, it is commonly reported in the literature to either merge multiple beats during processing \cite{chu2019ecg} or ensemble the predictions from multiple beats \cite{ihsanto2020fast}. 
Therefore, we have experimented with performing a majority voting over multiple consecutive beats to further improve the performance of EDITH. From the results presented in Fig. \ref{fig:cei_multi}, it can be observed that increasing the number of beats significantly improves the performance. We have limited our analysis up to 8 beats, as prior studies have presented this as an optimal trade-off between performance and practicality \cite{bonissi2013preliminary,odinaka2012ecg,labati2019deep}. In this regard, EDITH shows great promise by achieving the perfect $100\%$ accuracy using only 3 (ECG-ID, PTB) and 6 (MIT-BIH, NSRDB) beats. 

In Table \ref{tbl:cei}, we have presented a comparative analysis of closed environment identification performance of EDITH using the 60-20-20 split on the four different datasets, along with the \textit{state-of-the-art} results. Here, we only list the recent, top-performing deep learning-based approaches and it can be observed that EDITH outperforms them either by scoring higher or by achieving the same performance with less number of beats. 
%However, it should be noted that it is not possible to present a fair comparison of the methods, as different methods used different splits of data and some methods only reported results on validation data. Nevertheless, this analysis provides compelling evidence pointing to the potential of EDITH.

\begin{table}[h]
\tiny
\caption{Comparison with existing methods (Closed Environment Identification)}
\label{tbl:cei}
\centering
\begin{tabular}{|l|c|c|cV{4}l|c|c|c|}
\hline
\rowcolor[HTML]{9B9B9B} 
\multicolumn{4}{|cV{4}}{\cellcolor[HTML]{9B9B9B}ECG-ID} & \multicolumn{4}{c|}{\cellcolor[HTML]{9B9B9B}MIT-BIH} \\ \hline
\rowcolor[HTML]{C0C0C0} 
Method & \multicolumn{1}{l|}{\cellcolor[HTML]{C0C0C0}\begin{tabular}[c]{@{}l@{}}Number\\ of beats\end{tabular}} & \multicolumn{1}{c|}{\cellcolor[HTML]{C0C0C0}\begin{tabular}[c]{@{}c@{}}Evaluation\\ Scheme\end{tabular}} & \multicolumn{1}{lV{4}}{\cellcolor[HTML]{C0C0C0}Accuracy (\%)} & Method & \multicolumn{1}{l|}{\cellcolor[HTML]{C0C0C0}\begin{tabular}[c]{@{}l@{}}Number\\ of beats\end{tabular}} & \multicolumn{1}{c|}{\cellcolor[HTML]{C0C0C0}\begin{tabular}[c]{@{}c@{}}Evaluation\\ Scheme\end{tabular}} & \multicolumn{1}{l|}{\cellcolor[HTML]{C0C0C0}Accuracy (\%)} \\ \hline
\cite{salloum2017ecg} & 9 & Validation score & 100 & \cite{salloum2017ecg} & 9 & Validation score & 100 \\ \hline
\cite{ihsanto2020fast} & 8 & Validation score & 100 & \cite{ihsanto2020fast} & 6 & Validation score & 100 \\ \hline
\cite{chu2019ecg} & 2 & 10 Fold CV & 98.24 & \cite{chu2019ecg} & 2 & 10 Fold CV & 95.99 \\ \hline
\cite{lynn2019deep} & 9 & Test data & 98.60 & \cite{lynn2019deep} & 9 & Test data & 98.40 \\ \hline
\textbf{EDITH} & \textbf{3} & \textbf{Test data} & \textbf{100} & \textbf{EDITH} & \textbf{6} & \textbf{Test data} & \textbf{100} \\ \hline
\rowcolor[HTML]{9B9B9B} 
\multicolumn{4}{|cV{4}}{\cellcolor[HTML]{9B9B9B}PTB} & \multicolumn{4}{c|}{\cellcolor[HTML]{9B9B9B}NSRDB} \\ \hline
\rowcolor[HTML]{C0C0C0} 
Method & \begin{tabular}[c]{@{}c@{}}Number\\ of beats\end{tabular} & \begin{tabular}[c]{@{}c@{}}Evaluation\\ Scheme\end{tabular} & Accuracy (\%) & Method & \begin{tabular}[c]{@{}c@{}}Number\\ of beats\end{tabular} & \begin{tabular}[c]{@{}c@{}}Evaluation\\ Scheme\end{tabular} & Accuracy (\%) \\ \hline
\cite{chu2019ecg} & 2 & 10 Fold CV & 100 & \cite{chu2019ecg} & 2 & 10 Fold CV & 97.17 \\ \hline
\cite{labati2019deep} & 8 & Test data & 100 & \cite{zhang2017heartid} & 2-3 & Test data & 95.1 \\ \hline
\textbf{EDITH} & \textbf{3} & \textbf{Test data} & \textbf{100} & \textbf{EDITH} & \textbf{6} & \textbf{Test data} & \textbf{100} \\ \hline
\end{tabular}
\end{table}

\subsubsection{Cross-Session Evaluation}
For biometric authentication protocols, it is more meaningful to evaluate them using multi-session data, i.e., training the models with the data from a particular session and test them with data from a different session. Very few ECG databases provide us with multi-session data, ECG-ID and CYBHi are two of them. Therefore, we analyze the cross-session accuracy using these two datasets. We follow the similar evaluation criteria as followed by other works. We train the model using session 1 data and use the data from session 2 for evaluation, and vice versa. For this experiment, we just consider a single beat, in order to compare with other methods.

The result for cross-session analysis is presented in Table \ref{tbl:crs}. As it is intuitive and also reported by \cite{belo2020ecg}, identification accuracy sharply falls when tested on data from a different session, due to several variabilities. Despite that our model handles this quite well, as the accuracy on the ECG-ID dataset is still above 90\%. However BYCHi dataset being a more complex one, our performance falls below 80\%, still, it is over 10\% better than the previous best-performing work \cite{belo2020ecg}. For the ECG-ID dataset, although \cite{alduwaile2021using} performs better in such scenarios, their performance is not balanced, e.g. for some individuals, their model fails whereas our individual level accuracies are always over 70\%.

% \usepackage{multirow}
\begin{table}[]
\centering
\caption{Cross-Session Evaluation}
\label{tbl:crs}
\begin{tabular}{|c|c|c|c|}
\hline
Dataset & Method & \begin{tabular}[c]{@{}c@{}}Session 1\\ Accuracy (\%)\end{tabular} & \begin{tabular}[c]{@{}c@{}}Session 2\\ Accuracy (\%)\end{tabular} \\ \hline
\multirow{3}{*}{ECG-ID} & \cite{ihsanto2020fast} & 83.33 & - \\ \cline{2-4} 
 & \cite{alduwaile2021using} & 88.47-96.78 & 77.12-97.78 \\ \cline{2-4} 
 & EDITH & 91.67 & 92.671 \\ \hline
\multirow{2}{*}{CYBHi} & \cite{belo2020ecg} & 61 & 60.3 \\ \cline{2-4} 
 & EDITH & 71.597 & 73.929 \\ \hline
\end{tabular}
\end{table}

%\subsection{Siamese Architecture showed improved performance in the Identity Verification experiment}
\subsection{Identity Verification Experiments}

\subsubsection{Siamese Architecture vs Cosine Similarity}

\begin{figure*}[h]
    \centering
    \includegraphics[width=0.9\textwidth]{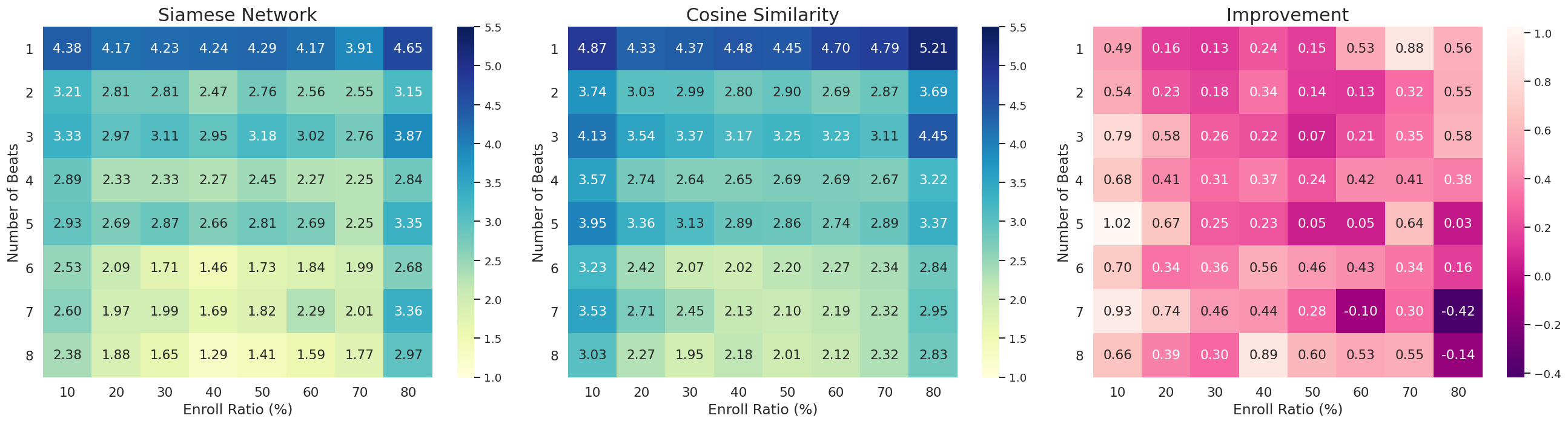}
    \caption{ Equal Error Rate (EER) obtained in different configurations for identity verification task. In addition to presenting the results for the proposed Siamese network, corresponding results using cosine similarity and the improvement in using Siamese network are also presented.}
    \label{fig:eer50}
\end{figure*}

As mentioned in the earlier sections, following the standard protocols to benchmark the identity verification task, we first train a model using a subset of individuals from the ECG-ID dataset and use that model to predict embeddings for the test signals. The test signals are further divided into two parts, enrollment and evaluation. It is common practice to keep 50\% data for enrollment and use the rest for evaluation \cite{chu2019ecg}. However, we experimented with different sizes of enrollment data and different numbers of beats. The EER values obtained for the different configurations are presented in Fig. \ref{fig:eer50}. In addition, we include the results using \textit{cosine similarity} and the resultant improvements achieved by using the Siamese network.

From Fig. \ref{fig:eer50}, we can observe that increasing the size of enrollment data at first gradually improves the performance, but then it starts to degrade. This can be explained as follows: if an enrollment set is limited, diversity in patterns may be insufficient (underfitting). If it is too large, this will enforce a bias on the matching task (overfitting). In our experiments, we have found that 40\% of the entire data used as the enrollment data yields the best performance. This is a reasonable setting, considering the prior works \cite{salloum2017ecg,chu2019ecg}, where 50\% data is used. On the other hand, increasing the number of beats almost consistently improves the performance. Again, it is evident that using the proposed Siamese network invariably yields better performance compared to the standard cosine similarity metric. %%%%Although the EER improvements may appear rather limited, it can be attributed to the superior feature learning capability of the proposed base model architecture. Since the model achieves an astounding accuracy in classifying the individuals in the closed environment, it must have captured the distinguishing features of the ECG signals. Consequently, even straightforward Cosine distance based on the model embedding is also able to distinguish the individuals. Having said that, the Siamese network enables us to further refine the results by weighing the degree of dissimilarity between the different features. 

In Table \ref{tbl:iv}, the results obtained by the comparative evaluation against the prior approaches are presented, where EDITH achieves the top performance, registering a better EER value than others.

\begin{table*}[h]
\centering
\caption{Comparison with existing methods for Identity Verification using the ECG-ID dataset.}
\label{tbl:iv}
\begin{tabular}{|
>{\columncolor[HTML]{C0C0C0}}c |c|c|c|c|c|}
\hline
Method & Zaghouani et al.\cite{zaghouani2017ecg} & Chun et al.\cite{chun2016single} & Chu et al.\cite{chu2019ecg} & Salloum et al.\cite{salloum2017ecg} & \textbf{EDITH} \\ \hline
EER & 15 \% & 5.2 \% & 2 \% & 1.7 \% & \textbf{1.29} \% \\ \hline
\end{tabular}
\end{table*}

\subsubsection{Case Study (Failure Scenarios)}

\begin{figure*}[h]
    \centering
    \includegraphics[width=0.75\textwidth]{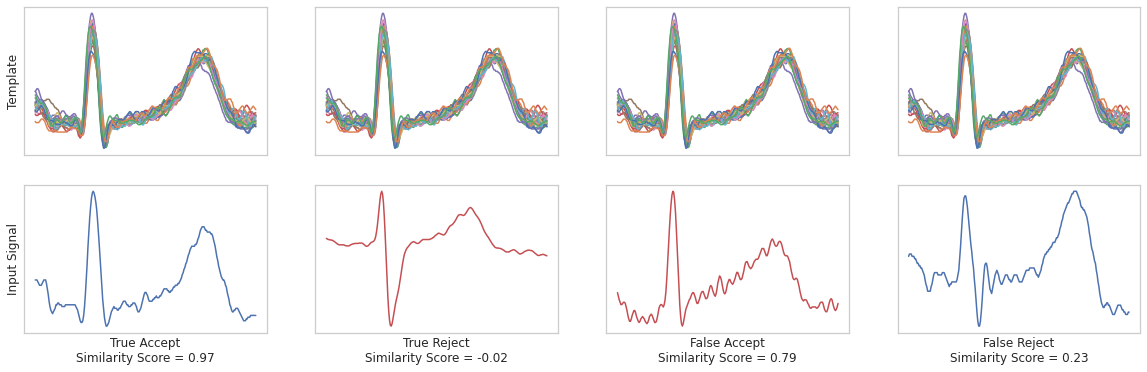}
    \caption{Example cases of identity verification. Here, we can observe that signals from the same (blue) /different (red) individuals, match/mismatches with the templates. However, due to the variability in ECG signal, sometimes signal from a different person may match with the templates (false accept), similarly signal from the same person may deviate too much from the templates (false reject).}
    \label{fig:compare}
\end{figure*}

We further analyze some cases where EDITH failed in verifying the identities. Fig. \ref{fig:compare} presents some relevant examples. For the true acceptance and true rejection cases, the reasons for the correct decision are clearly visible (negligible S region for true accept case vs. prominent S region for true reject case) through visual comparison of the input with the template. However, for some cases, signals from other individuals closely mimicked the pattern of the templates (false accept). On the contrary, in some situations, the ECG signal from the same person may deviate significantly, and hence, it hardly matches the template (salient S region in false reject). The level of variability and inconsistency in the ECG signals are responsible for such erroneous decisions, this is the primary motivation for analyzing multiple beats instead of a single one.

\subsubsection{Result on Other Datasets}

\begin{figure}[h]
     \centering
     \begin{subfigure}[b]{0.24\textwidth}
         \centering
         \includegraphics[width=\textwidth]{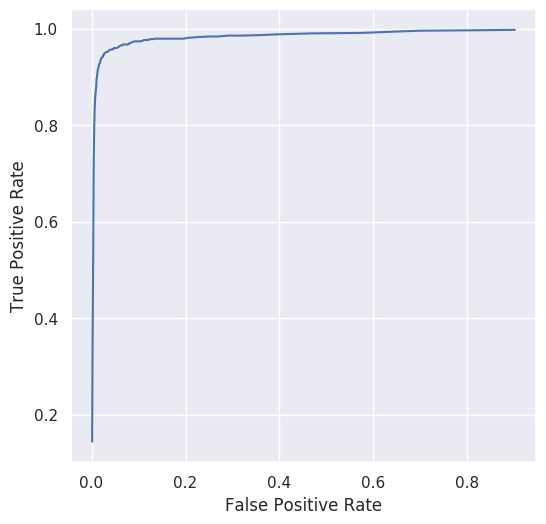}
         \caption{ECG-ID\\(AUC = 94.41\%)}
         \label{fig:roc_ecgid}
     \end{subfigure}
     \hfill
     \begin{subfigure}[b]{0.24\textwidth}
         \centering
         \includegraphics[width=\textwidth]{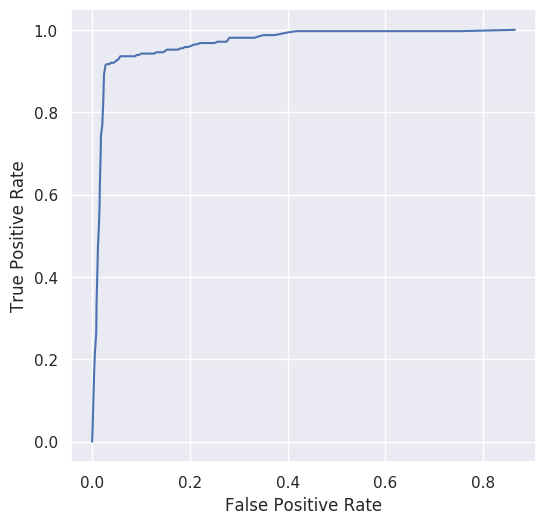}
         \caption{MIT-BIH\\(AUC = 87.76\%)}
         \label{fig:roc_mitbih}
     \end{subfigure}
     \hfill
     \begin{subfigure}[b]{0.24\textwidth}
         \centering
         \includegraphics[width=\textwidth]{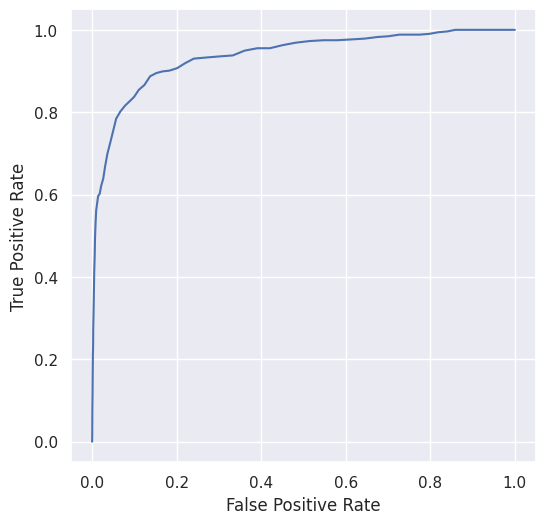}
         \caption{PTB\\(AUC = 84.88\%)}
         \label{fig:ptb_roc}
     \end{subfigure}
     \hfill
     \begin{subfigure}[b]{0.24\textwidth}
         \centering
         \includegraphics[width=\textwidth]{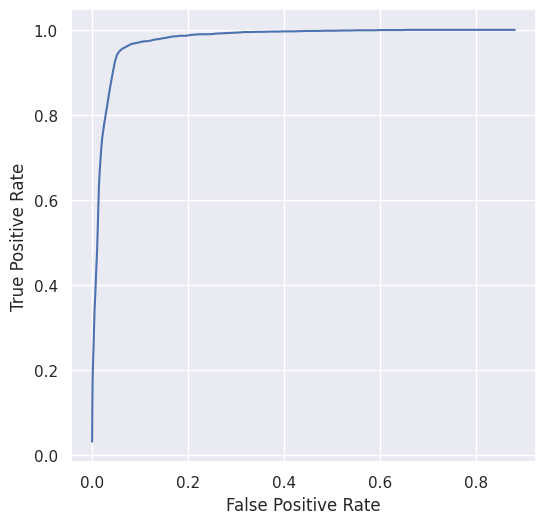}
         \caption{NSRDB\\(AUC = 86.29\%)}
         \label{fig:roc_nsr}
     \end{subfigure}
        \caption{ROC Curves of Identity Verification on Different Datasets.}
        \label{fig:roc}
\end{figure}

We also evaluated our identity verification performance on the other datasets. Since very few works were evaluated on datasets other than ECG-ID, we could not present any meaningful comparison. We present the EER and ROC curves in Fig. \ref{fig:eer} and Fig. \ref{fig:roc} respectively. The results here presented are for single beat only with 50\% enrollment rate. The scores on other datasets are less than ECG-ID dataset because the model was adapted for ECG-ID dataset and just fine-tuned on the other datasets.

\begin{figure}[h]
     \centering
     \begin{subfigure}[b]{0.24\textwidth}
         \centering
         \includegraphics[width=\textwidth]{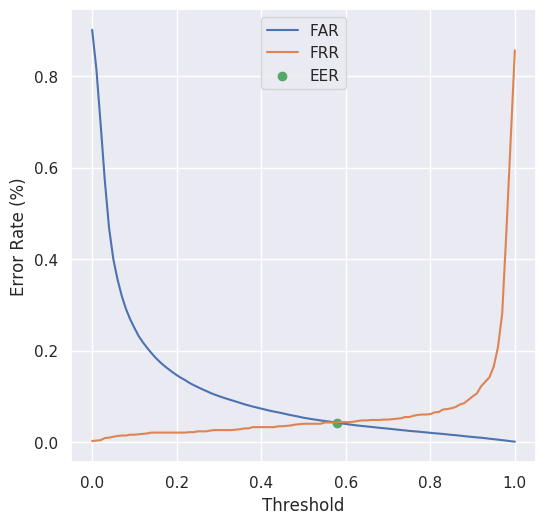}
         \caption{ECG-ID (EER = 4.27\%)}
         \label{fig:eer_ecgid}
     \end{subfigure}
     \hfill
     \begin{subfigure}[b]{0.24\textwidth}
         \centering
         \includegraphics[width=\textwidth]{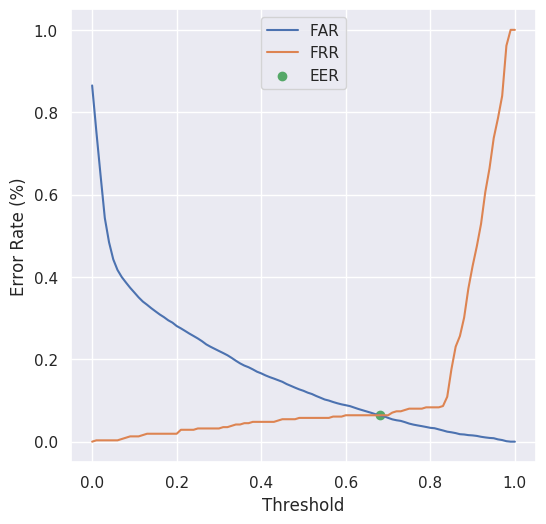}
         \caption{MIT-BIH (EER = 6.36\%)}
         \label{fig:eer_mitbih}
     \end{subfigure}
     \hfill
     \begin{subfigure}[b]{0.24\textwidth}
         \centering
         \includegraphics[width=\textwidth]{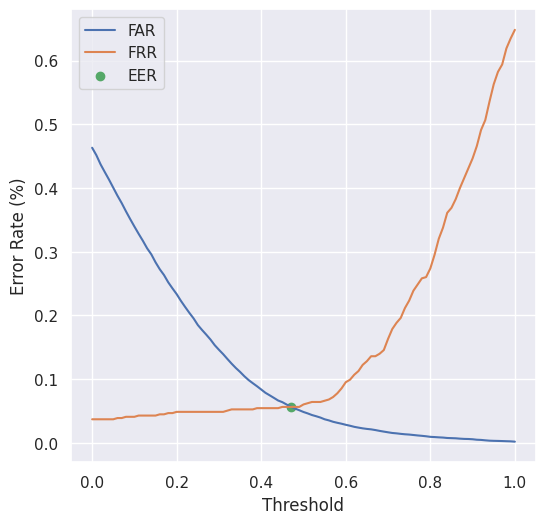}
         \caption{PTB (EER = 5.66\%)}
         \label{fig:ptb_eer}
     \end{subfigure}
     \hfill
     \begin{subfigure}[b]{0.24\textwidth}
         \centering
         \includegraphics[width=\textwidth]{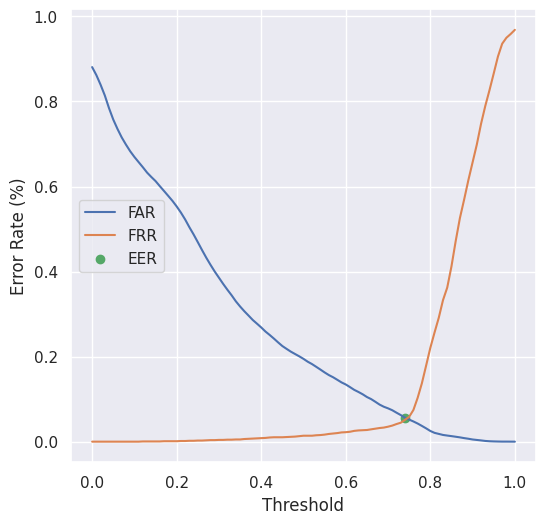}
         \caption{NSRDB (EER = 5.17\%)}
         \label{fig:eer_nsr}
     \end{subfigure}
        \caption{EER Curves of Identity Verification on Different Datasets.}
        \label{fig:eer}
\end{figure}

\subsection{Investigation on model interpretability}

The improved performance of machine learning often comes at the cost of interpretability. However, for security and biometrics applications, interpretation of algorithms is imperative. Not only this will enable us to assert that the model is learning actual significant information, but also it will act as a defense mechanism against spoofing or counterfeiting attacks. Recently, considerable focus has been put on interpreting the deep learning models, instead of considering them merely as black-boxes \cite{gilpin2018explaining}.

\begin{figure}[h]
     \centering
     \begin{subfigure}[b]{0.2\textwidth}
         \centering
         \includegraphics[width=\textwidth]{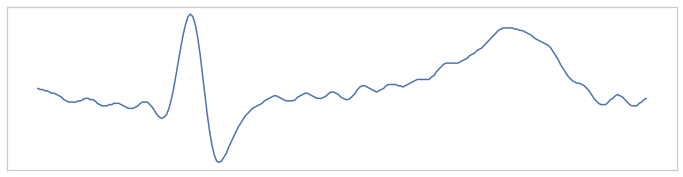}
         \caption{}
         \label{fig:intrprt_signal}
     \end{subfigure}
     \hfill
     \begin{subfigure}[b]{0.79\textwidth}
         \centering
         \includegraphics[width=\textwidth]{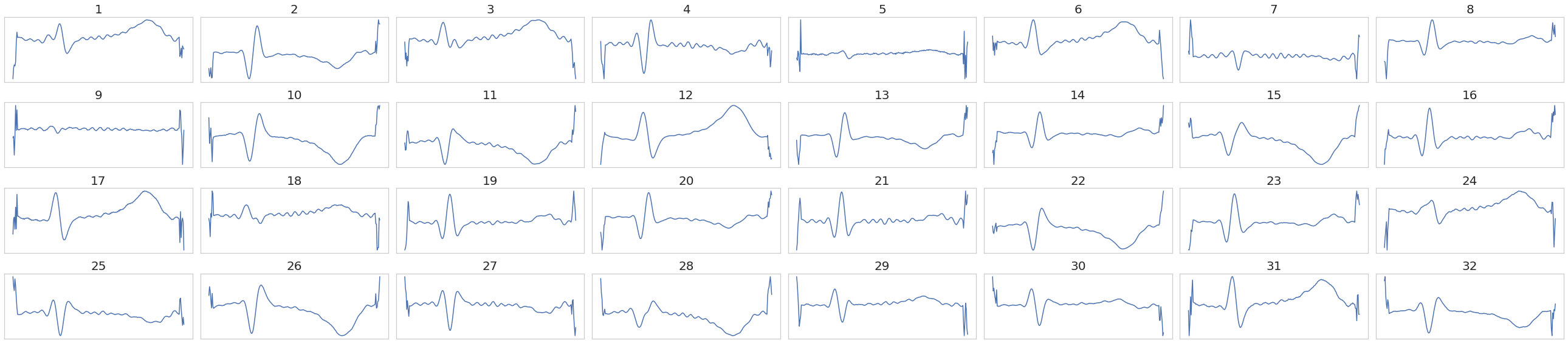}
         \caption{}
         \label{fig:intrprt_fitler}
     \end{subfigure}
        \caption{Demonstration of the features learned by the filters. For a random signal from the ECG-ID database (\ref{fig:intrprt_signal}), we visualize the output featuremaps of the filters from the first convolutional layer (\ref{fig:intrprt_fitler}). Some filters have identified the R (19), S (7), T (24) regions, whereas some just denoise (12) or invert (26) the signal.}
        \label{fig:intrprt}
\end{figure}

With the above backdrop, we investigate the interpretability of our proposed method. First of all, to analyze the discriminative features extracted by the model for a random signal from the ECG-ID database (Fig. \ref{fig:intrprt_signal}), we visualize the feature maps computed by the first convolutional layer (Fig. \ref{fig:intrprt_fitler}). Although the earlier feature maps alone do not sufficiently signify, still we can obtain some rough ideas about the features. For example, it can be observed that some filter seeks for the R (19), S (7), T (24) regions, while some filter just merely denoise (12) or invert (26) the signal.

Saliency maps \cite{simonyan2013deep} provide further insights into the interpretability  of the models, which highlight the parts of the signal that influence the most in making the predictions. Saliency maps can be generated by computing the gradient of the loss function with respect to the individual timestamps of the input signal \cite{shen2019ambulatory}. For example, for an $n$ sample long input signal, $X=[x_1,x_2,x_3,\dots,x_n]$, we can compute the saliency map, $S=[s_1,s_2,s_3,\dots,s_n]$ as $s_i=|\frac{\delta L}{\delta x_i}|,\forall i \epsilon [1 \dots n]$, where $L$ is the value of the loss function.

\begin{figure}[h]
    \centering
    \includegraphics[width=\textwidth]{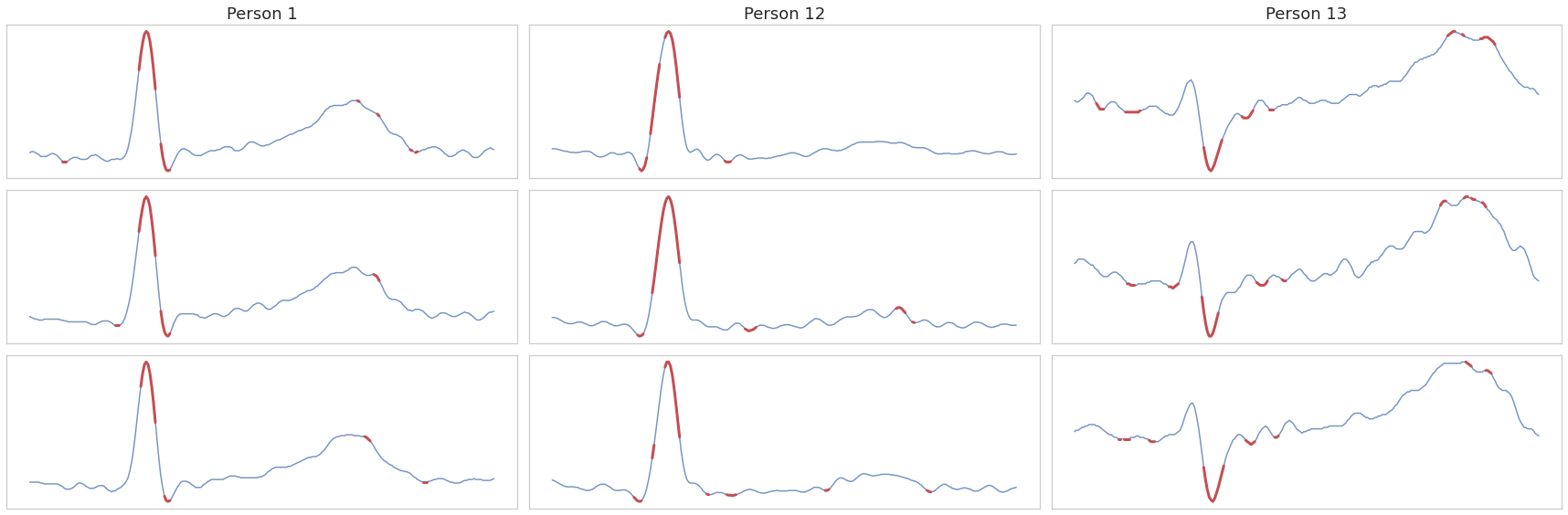}
    \caption{Saliency maps for randomly selected 3 different persons from the ECG-ID database. It is interesting to observe that the model has learned the fiducial landmarks like Q,R,S sub-waves, without any explicit supervision.}
    \label{fig:intrprt_saliency}
\end{figure}

Fig. \ref{fig:intrprt_saliency} presents the saliency maps of 3 different signals from 3 different persons from the ECG-ID database, where the regions with high saliency scores are colored in red. For Person 1, we have a clear S region and vague Q region, but for Person 12 we observe the opposite. Thus, in order to discriminate them, one would look for the Q and S regions, which are the ones sought by the model as demonstrated in Fig. \ref{fig:intrprt_saliency}. Since signals from both the person have a conspicuous R peak, the model emphasizes that as well. However, for Person 13, the R peak is less dominant, thus the model has put little emphasis there. This analysis indicates the model's capability of identifying the intelligible fiducial landmarks and making the prediction accordingly.

\subsection{Evaluation with Real-World Data}

In order to assess the practical utility of the proposed method, we also perform a limited trial. We collect ECG signals from 9 persons using HealthyPi v4\footnote{https://healthypi.protocentral.com/}. This kit contains an ECG front end of Texas Instruments (TI) ADS1292R with 24-bit resolution, having a signal-to-noise ratio (SNR) of 107 dB. The ECG signals are recorded at 125 Hz, which are upsampled to 500 Hz using bicubic interpolation for our analysis.

\begin{figure}[h]
     \centering
     \begin{subfigure}[b]{0.49\textwidth}
         \centering
         \includegraphics[width=\textwidth]{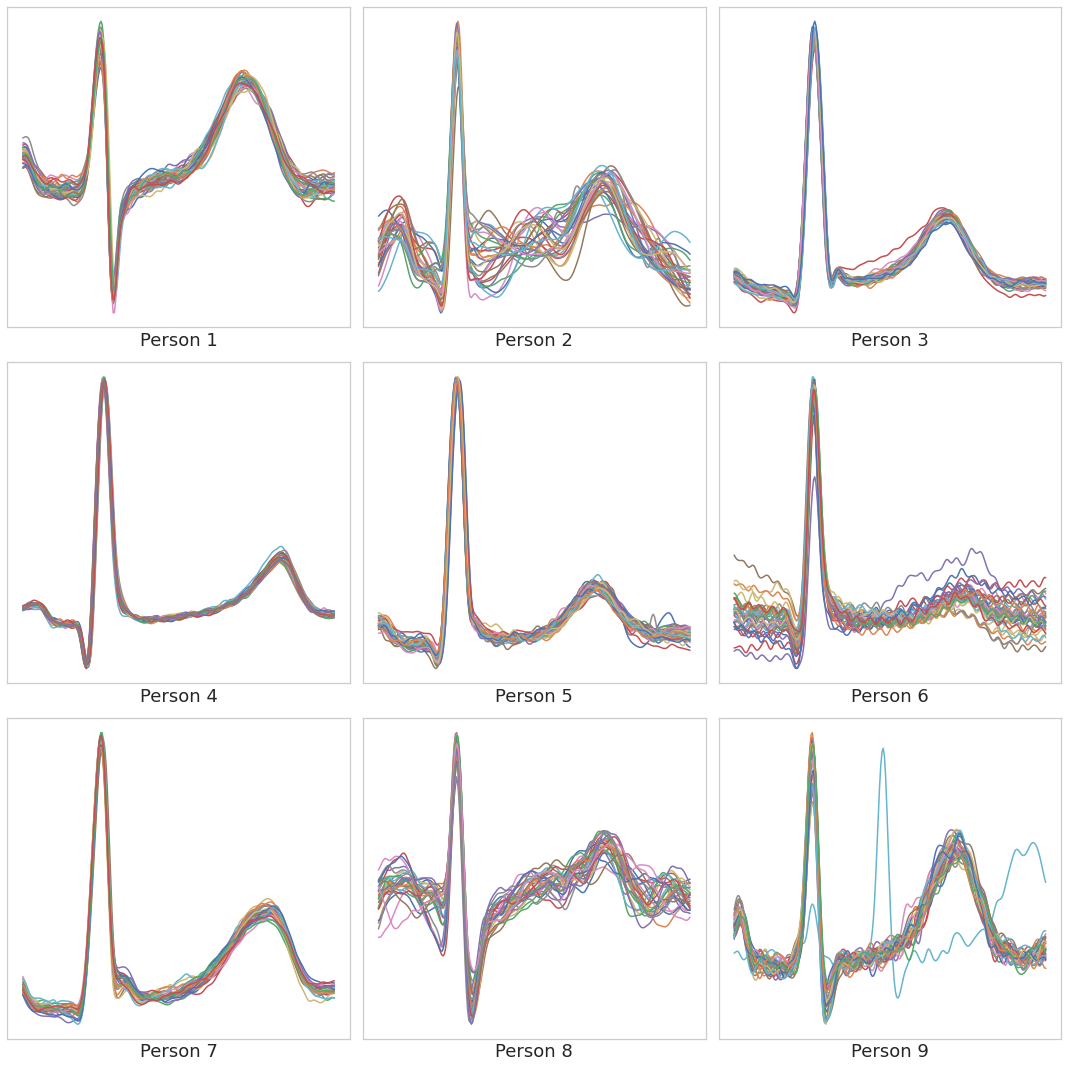}
         \caption{}
         \label{fig:qu_rpeak}
     \end{subfigure}
     \hfill
     \begin{subfigure}[b]{0.49\textwidth}
         \centering
         \includegraphics[width=\textwidth]{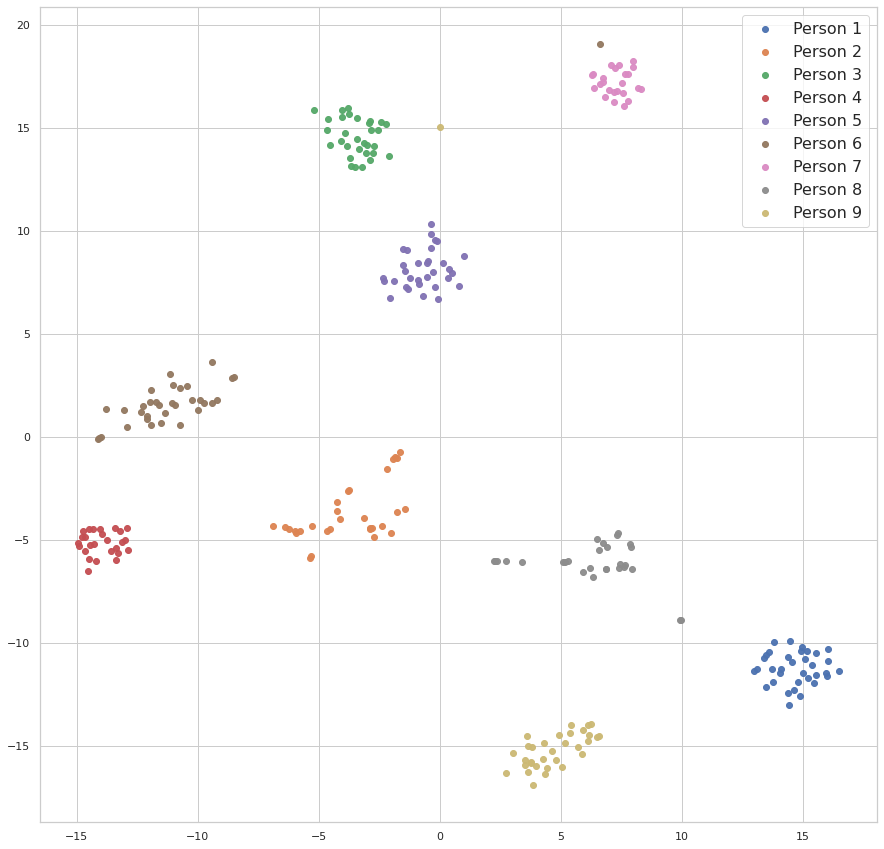}
         \caption{}
         \label{fig:qu_tsne}
     \end{subfigure}
        \caption{Evaluation with real-world data.  (\ref{fig:qu_rpeak}) illustration of reliable R-peak detection
        and (\ref{fig:qu_tsne}) the embeddings for different individuals obtained using the closed environment identification model.}
        \label{fig:qudata}
\end{figure}

We take the R-peak detector model and the closed environment identification model trained on the ECG-ID database and investigate their applicability with this completely different source of data. The model successfully detects the R-peaks, leading to a proper segmentation of the beats. In Fig. \ref{fig:qu_rpeak} we have presented an overlapping plot of the ECG beats for different persons and it can be observed that all the beats have been properly segmented except one.

Similarly, we pass the detected beats to the identification model and compute the embeddings. We present a 2-dimensional t-distributed stochastic neighbor embedding (TSNE) \cite{maaten2008visualizing} plot of the embeddings in Fig. \ref{fig:qu_tsne}. It can be observed that the embeddings from individual persons are clustered together, which indicates that a basic fine-tuning or template matching would be able to identify the persons.

Therefore, this limited study demonstrates the potential of EDITH in real-world authentication applications. It is indeed promising that the model was able to generalize to this extent, despite the data being captured using a different device through a different protocol, and that too with a different sampling rate and quantization.

To sum up, owing to the continual adversarial attacks and forgery attempts, security researchers are in constant pursuit of more reliable and robust biometric authentication systems. Leveraging physiological signals seems to be the next pivotal step in this regard, due to the inherent defense mechanism thereof against forgery. The proposed EDITH system is a step towards attaining this objective. We have demonstrated superior performance of EDITH over the existing methods in different tasks using various datasets. Furthermore, EDITH circumvents certain limitations of the contemporary methods, for instance, inconsistent R-peak detection and the need for longer ECG signals.

\section{Conclusion}

The domains of security and biometrics have always been an arms race. As efforts were put in developing more advanced authentication systems, a similar level of striving (or perhaps, even more) was in place to forge them. For example, 3D facial biometrics was introduced to overcome the limitations of 2D face scanning biometrics, as mere images were able to spoof them. But still, those systems were still vulnerable to 3D face model based counterfeiting attacks \cite{jia2019database}. As a result, researchers from this domain have not only been attempting to improve the existing systems but also are constantly looking for more reliable alternatives. In this regard, physiological signal based biometrics, bring in greater promises as they are comparatively difficult to fake.

In this work, we have presented EDITH, a biometric authentication system based on ECG signals and deep learning techniques. Similar to \cite{ihsanto2020fast} EDITH was developed to work with a single heartbeat, but its performance can be enhanced by merging multiple beats. We propose a novel architecture leveraging multiresolution analysis. As a result, EDITH outperforms existing methods by achieving better accuracy with less number of beats. Furthermore, we present a competitive alternative to the widely adopted \textit{Cosine similarity}-based identity verification. The proposed novel Siamese architecture, incorporating both Euclidean and Cosine-like distances consistently surpasses Cosine similarity-based template matching. Another remarkable achievement of EDITH lies in its ability to learn the fiducial points of ECG signals without any explicit supervision. Furthermore, a limited trial with real-world ECG data corroborates the potential of EDITH as a practical authentication system.

Perhaps, physiological signal based biometrics are yet to be sufficient for real-world deployment. Nevertheless, for the favorable characteristics they bring in, such means of authentication is worth investigating further. The future directions of this research can be manifold. It is indeed true that owing to the variability and diversity using a single heartbeat is unlikely to offer reliable authentication. Despite this, it deserves further investigation to attain the holy grail of single beat authentication. Analysis of variation in ECG signals with time along with different states of the individual is another avenue that deserves more attention to make authentication more robust and steadfast.

\section*{Acknowledgement}
This work was supported by Grant NPRP12S-0227-190164 from the Qatar National Research Fund, a member of Qatar Foundation, Doha, Qatar and the claims made herein are solely the responsibility of the authors. Open Access funding provided by the Qatar National Library.

\section*{Competing interests}
The authors declare no competing interests.

\section*{Author Contributions}
Nabil Ibtehaz: Conceptualization, Methodology, Software, Validation, Formal analysis, Writing - Original Draft, Writing - Review \& Editing.
Muhammad E. H. Chowdhury: Conceptualization, Writing - Original Draft, Writing - Review \& Editing, Supervision, Project administration.
Amith Khandakar: Data Curation, Methodology, Visualization, Resources, Writing - Review \& Editing. 
Serkan Kiranyaz: Conceptualization, Supervision, Writing - Review \& Editing.
M. Sohel Rahman: Conceptualization, Supervision, Writing - Review \& Editing.
Anas Tahir: Data Curation, Investigation, Resources, Writing - Review \& Editing.
Yazan Qiblawey: Data Curation, Investigation, Resources, Writing - Review \& Editing.
Tawsifur Rahman: Methodology, Formal analysis, Writing - Review \& Editing.
All the authors reviewed, and revised the draft and approved the final version of the article before submission.

\bibliographystyle{unsrt}
\bibliography{main.bib}

\end{document}